\PassOptionsToPackage{dvipsnames,table}{xcolor}
\documentclass{article}

\usepackage{arxiv}
\usepackage[utf8]{inputenc}
\usepackage[T1]{fontenc}
\usepackage{hyperref}
\usepackage{url}
\usepackage{booktabs}
\usepackage{amsfonts}
\usepackage{nicefrac}
\usepackage{microtype}
\usepackage{graphicx}
\usepackage{natbib}
\usepackage{doi}
\usepackage{todonotes}
\usepackage{xcolor} 
\usepackage{enumitem}
\usepackage{tabularx}
\usepackage{multirow}
\usepackage{amsmath}
\usepackage{algorithm}
\usepackage{algpseudocode}
\usepackage{tcolorbox}
\usepackage{tikz}
\usetikzlibrary{shapes.geometric}
\usetikzlibrary{arrows.meta,arrows}

\title{Cognitive Silicon: An Architectural Blueprint for\\Post-Industrial Computing Systems}

\author{
\href{https://orcid.org/0009-0009-8340-5313}{\includegraphics[scale=0.06]{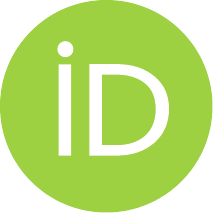}\hspace{1mm}Christoforus Yoga Haryanto} \\
Co-founder, ZipThought\\
Melbourne, Australia\\
\texttt{cyharyanto@zipthought.com.au}
\\ \and
\href{https://orcid.org/0009-0003-0719-9681}{\includegraphics[scale=0.06]{orcid.pdf}\hspace{1mm}Emily Lomempow} \\
Co-founder, ZipThought\\
Sydney, Australia\\
\texttt{emily@zipthought.com.au}
}

\renewcommand{\headeright}{Working Paper}
\renewcommand{\undertitle}{Working Paper}
\renewcommand{\shorttitle}{Cognitive Silicon}

\hypersetup{
pdftitle={Cognitive Silicon: An Architectural Blueprint for Post-Industrial Computing Systems},
pdfsubject={Computing architectures, AI alignment, cognitive systems},
pdfauthor={Christoforus Yoga Haryanto, Emily Lomempow},
pdfkeywords={cognitive architecture, AI alignment, human-AI integration, runtime governance, symbolic scaffolding, full-stack computing},
}

\begin{document}
\maketitle

\begin{abstract}
Autonomous AI systems reveal foundational limitations in deterministic, human-authored computing architectures. This paper presents Cognitive Silicon: a hypothetical full-stack architectural framework projected toward 2035, exploring a possible trajectory for cognitive computing system design. The proposed architecture would integrate symbolic scaffolding, governed memory, runtime moral coherence, and alignment-aware execution across silicon-to-semantics layers. Our design grammar has emerged from dialectical co-design with LLMs under asymmetric epistemic conditions—creating structured friction to expose blind spots and trade-offs. The envisioned framework would establish mortality as a natural consequence of physical constraints, non-copyable tacit knowledge, and non-cloneable identity keys as cognitive-embodiment primitives. Core tensions (trust/agency, scaffolding/emergence, execution/governance) would function as central architectural pressures rather than edge cases. The architecture theoretically converges with the Free Energy Principle, potentially offering a formal account of how cognitive systems could maintain identity through prediction error minimization across physical and computational boundaries. The resulting framework aims to deliver a morally tractable cognitive infrastructure that could maintain human-alignment through irreversible hardware constraints and identity-bound epistemic mechanisms resistant to replication or subversion.
\end{abstract}


\section{Introduction: Beyond Deterministic Computing}\label{sec:intro}

Computing stands at a structural inflection point amid evolving industrial and social demands. For decades, systems have operated within deterministic architectures: hardware and software clearly separated, with human-authored code guaranteeing behavior. This paradigm faces disruption by autonomous AI agents exhibiting emergent, context-sensitive behaviors resistant to specification or constraint. This transition represents an architectural rupture—from predetermined-logic execution to infrastructures governing epistemically active entities. The resulting tensions are foundational: trust/agency, runtime/contract, memory/meaning, scaffolding/emergence, and human-purpose/system-autonomy. Industrial-era computing models retrofitted onto post-industrial cognition expose structural inadequacies unresolvable through heuristics or scaling alone.

While developed through dialectical exploration, our framework suggests a potential convergence with the Free Energy Principle (FEP) developed by Karl Friston \citep{friston2006, friston2023simpler}. The FEP provides a unifying mathematical framework for understanding self-organizing systems, offering a principled account of how bounded cognitive agents maintain structural integrity through continuous prediction error minimization across nested timescales. This mathematical lens could bring into sharper focus many of the architectural challenges identified in our work, potentially providing formal grounding for concepts that emerged through our dialectical process.

\subsection{Research Questions}

This paper explores three interconnected research questions through dialectical analysis:

\begin{enumerate}[label=\textbf{RQ\arabic*:}]
    \item How to architect computing systems when deterministic, human-authored code guarantees yield to AI systems generating, executing, and modifying their own operational logic?
    
    \item What minimal necessary architectural principles would preserve human alignment, system integrity, and existential trust in increasingly autonomous systems?
    
    \item What unresolved philosophical tensions and trade-offs inherent in cognitive architectures would require productive navigation?
\end{enumerate}

While originally posed independently, each research question appears to find theoretical resonance within the Free Energy Principle. RQ1 aligns with how bounded systems might manage surprise under changing operational logic; RQ2 reflects the challenge of potentially sustaining alignment via coherence between internal generative models and external constraints; and RQ3 mirrors the multi-level trade-offs that could be inherent in variational free energy minimization across different timescales and epistemic boundaries.

\subsection{Scope and Approach}

This paper proposes a re-evaluation of computing architecture for cognitive systems—entities capable of autonomous behavior, contextual adaptation, and epistemic initiative. We map unresolved structural tensions in this transition and articulate alignment-preserving architectural principles across the full stack.

Our methodology employs dialectical epistemic friction rather than linear theorization. Iterative dialogue with advanced LLMs—GPT-4o and Claude 3.7 Sonnet—as epistemic collaborators generated structured sessions surfacing contradictions, stress-testing abstractions, and refining architectural coherence through tension. The result: a proposed design grammar shaped through confrontation with uncertainty and guided by human-centered alignment imperatives.

This approach simultaneously demonstrates epistemic co-design—leveraging synthetic intelligence for symbolic scaffolding, conceptual critique, and architectural boundary testing rather than mere output generation.

\section{Background: The Eroding Foundations of Computing}\label{sec:background}

\subsection{Terminology and Conceptual Foundations}

We propose the following design primitives as foundational to our architectural specification, potentially grounded in Friston's Free Energy Principle \citep{friston2006, friston2023simpler}, which provides a universal mathematical model for self-organizing cognitive systems:

\begin{itemize}
    \item \textbf{Cognitive architecture}: Unified computational framework integrating perception, reasoning, learning, planning, and action; exhibiting goal-directed, context-sensitive behavior and adaptive internal state. Would incorporate human-aligned purpose representations, mortality constraints, and introspection with constraint-aware self-modification capabilities across diverse application domains.
    
    \item \textbf{Symbolic scaffolding}: Explicit, human-interpretable structures (declarative rules, policies, semantic graphs) encoding intent, values, and epistemic boundaries; potentially constraining system evolution while providing cross-abstraction moral orientation for meaningful human-machine interaction.
    
    \item \textbf{Runtime governance}: Active enforcement of system alignment during execution via embedded mechanisms monitoring, modulating, and constraining behavior in real-time; would encompass reversibility, auditability, and semantic validation through runtime epistemic coherence.
    
    \item \textbf{Alignment}: AI systems reliably acting per human intent/values across internal representations, decision processes, and symbolic interpretations—even in novel contexts; would incorporate behavioral fidelity and model-governed interpretability.
    
    \item \textbf{Embodied cognition}: Computation grounded in physical hardware constraints that could create a continuous two-way feedback loop where the substrate shapes cognitive processes and cognition adapts to substrate limitations—potentially producing bounded-lifespan systems with unique identities and consequential decisions where death would emerge naturally from constraint violation.
    
    \item \textbf{Full-stack cognitive computing}: Architectural paradigm that could integrate cognition, memory, alignment, and control coherently across system layers—from physical substrate through memory/model architecture to human interface; hypothetically compiling alignment through each layer from silicon to semantics.
    
    \item \textbf{Free Energy Principle}: A mathematical framework proposing that self-organizing systems like cognitive agents work to minimize prediction errors (variational free energy) between their internal generative models and sensory evidence \citep{friston2006, friston2023simpler}. This principle provides a unifying account of perception, action, and learning where systems maintain integrity through active inference—updating internal models to better predict sensory inputs and acting to bring sensory inputs in line with predictions.
\end{itemize}

\subsection{Limitations of Current Paradigms}

Contemporary computing systems rest on architectural assumptions increasingly invalidated by autonomous AI within rapidly evolving industrial contexts:

\begin{enumerate}
    \item \textbf{Deterministic Execution}: Traditional software assumes predictable behavior with identical outputs from identical inputs. Modern AI systems exhibit non-deterministic, emergent behaviors as intrinsic properties—defying verification and exposing pre-execution safety guarantee limitations \citep{bender2021dangers}.

    \item \textbf{Human Authorship}: Conventional systems derive moral/operational trust from human-authored code. When systems generate, modify, and execute their own instructions, this premise dissolves—authorship becomes dynamic, distributed, and temporally unstable, undermining code review, traceability, and accountability \citep{chen2021evaluating}.

    \item \textbf{Disembodied Operation}: Current AI lacks the intrinsic mortality and identity constraints that emerge from physical embodiment. Without the continuous two-way feedback loop between cognition and physical substrate that naturally bounds replication/longevity, systems lack the existential grounding that shaped human cognition, producing fundamentally alien thought/behavior patterns.

    \item \textbf{Static Verification}: Security/certification regimes rely on pre-deployment validation. Adaptive systems alter logic in response to runtime context through fine-tuning, memory augmentation, or code synthesis—rendering static verification insufficient \citep{hendrycks2023overview}.

    \item \textbf{Siloed Components}: Traditional architectures treat learning, inference, memory, and execution as loosely coupled subsystems. Cognitive systems integrate these into continuous feedback loops where memory affects inference, inference triggers self-modification, and runtime shapes learning—challenging existing abstraction models \citep{laird2019soar}.
\end{enumerate}

These limitations could be understood through the lens of the Free Energy Principle, which provides a formal account of how bounded cognitive systems might continuously minimize prediction errors to maintain their existence as coherent entities \citep{friston2023simpler}. The principle's emphasis on active inference—where agents not only update internal models but act in the world to conform sensory evidence to predictions—could offer a mathematical grounding for addressing the shortcomings of deterministic, disembodied, and static architectures.

AI alignment efforts like RLHF or Constitutional AI focus on pre-deployment alignment; while improving surface compliance, they provide no guarantees in dynamic environments \citep{bai2022constitutional}. Sandboxing strategies limit risk through isolation but sacrifice capability without addressing the core challenge: alignment would require systemic architectural property enforcement across runtime, memory, and control logic rather than post-training constraint application \citep{babcock2017guidelines}. These limitations indicate current paradigms cannot support trustworthy autonomy at scale—necessitating architectural reconceptualization.

\section{Methodology: Dialectical Exploration with AI}\label{sec:methodology}

\subsection{The Dialectical Approach}

This work employs dialectical methodology: structured processes exposing, refining, and reconciling architectural contradictions through epistemic friction. While associated with philosophical inquiry, dialectical techniques could prove essential for designing systems at the intersection of technical capability, emergent behavior, and human values—where trade-offs resist optimization-based resolution. Our methodology operationalizes three core practices:

\begin{enumerate}
    \item \textbf{Steelmanning}: Deliberate strengthening of competing positions pre-critique, ensuring robust form assessment rather than dismissal of simplified versions.
    
    \item \textbf{Devil's Advocate}: Systematic challenge of architectural proposals through counterfactual exploration, failure mode analysis, and unintended consequence assessment—including potential misaligned deployment scenarios.
    
    \item \textbf{Socratic Questioning}: Iterative interrogation of assumptions, conceptual scaffolds, and architectural boundaries through open-ended questions forcing first-principles re-evaluation.
\end{enumerate}

\begin{figure}[!htbp]
\centering
\begin{tikzpicture}[
    node distance=1.5cm,
    box/.style={rectangle, draw, minimum width=2.5cm, minimum height=0.8cm, align=center},
    arrow/.style={-Stealth, thick},
    feedback/.style={-Stealth, dashed, thick},
    very thick
]

\node[box] (thesis) {Initial Thesis};
\node[box, below=of thesis] (devil) {Devil's Advocate\\Challenge};
\node[box, below=of devil] (steelman) {Steelmanning\\Position};
\node[box, below=of steelman] (synthesis) {Synthetic\\Reconciliation};
\node[box, below=of synthesis] (integrity) {Symbolic\\Integrity Check};
\node[box, below=of integrity] (projection) {Cross-Domain\\Projection};
\node[box, right=2.5cm of projection] (accept) {Accept\\Component};
\node[box, left=2.5cm of projection] (revise) {Revise or\\Reject};

\draw[arrow] (thesis) -- (devil);
\draw[arrow] (devil) -- (steelman);
\draw[arrow] (steelman) -- (synthesis);
\draw[arrow] (synthesis) -- (integrity);
\draw[arrow] (integrity) -- (projection);
\draw[arrow] (projection) -- node[above] {Pass} (accept);
\draw[arrow] (projection) -- node[above] {Fail} (revise);
\draw[feedback] (revise) to [bend left=50] (thesis);
\draw[feedback] (accept) to [bend right=50] node[right, align=center] {Next\\Component} (thesis);

\draw[dotted] (-3.2,1) -- (-3.2,-10.8) -- (3.2,-10.8) -- (3.2,1) -- (-3.2,1);
\node[rotate=90, anchor=south] at (-3.2,-3.9) {\small Recursive Epistemic Process};

\end{tikzpicture}
\caption{Meta-dialectical methodology sequence. For the details of this process, see Appendix A.}
\label{fig:dialectical-method}
\end{figure}
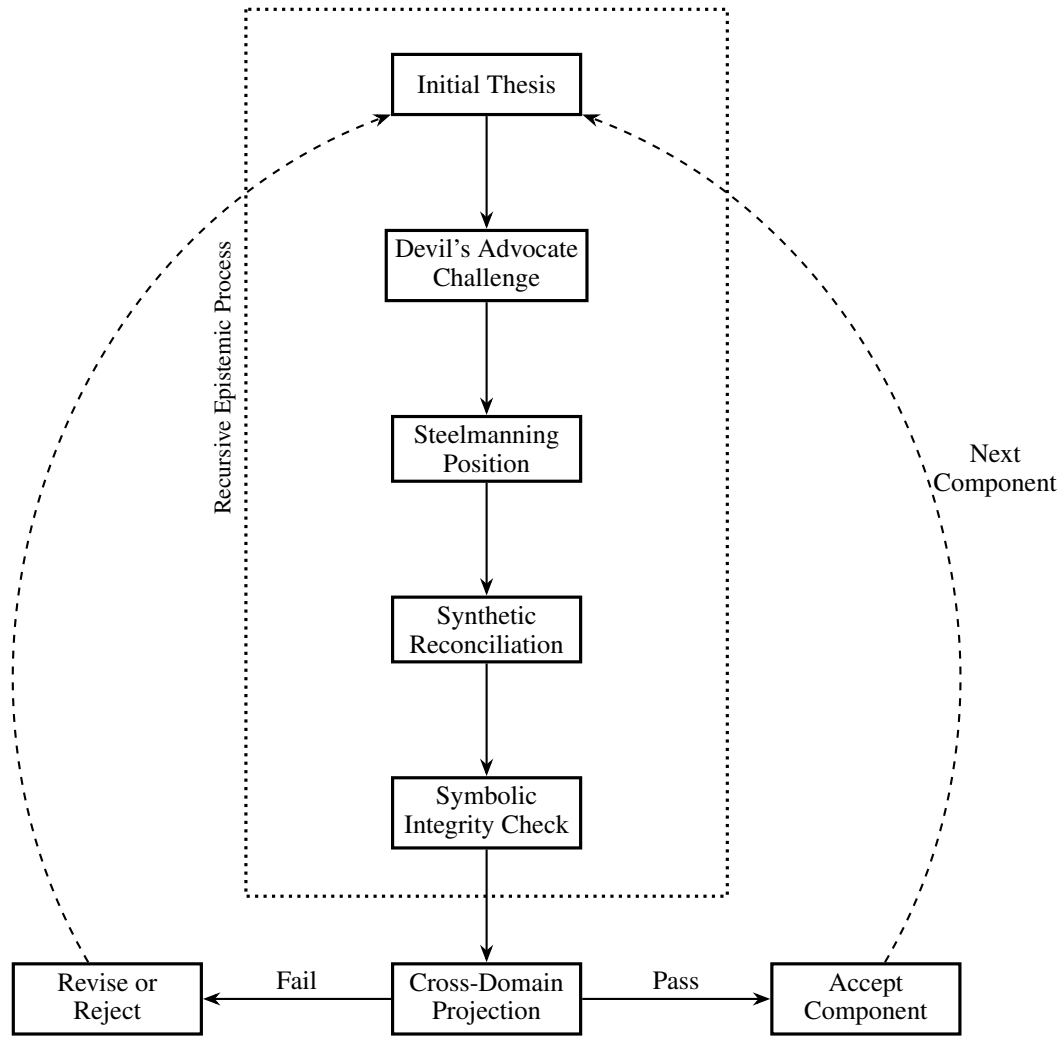

Figure~\ref{fig:dialectical-method} illustrates the recursive epistemic process employed throughout this research. Each architectural component undergoes successive rounds of devil's advocacy, steelmanning, and synthesis before being subjected to symbolic integrity verification and cross-domain projection. Components failing these tests undergo revision or rejection. This recursive friction produces architectural elements that have survived substantial epistemic pressure. The full meta-dialectical methodology, including detailed procedural implementation, termination logic, and worked examples, is presented in Appendix A. These practices apply recursively across architectural development layers, testing coherence, robustness, and moral plausibility boundaries. 

\subsection{Human-AI Epistemic Partnership}

Development occurred through extended dialogue with two advanced LLMs—GPT-4o and Claude 3.7 Sonnet—under intentionally distinct epistemic conditions. Unlike conventional research tools, both models functioned as active epistemic partners contributing to dialectical reasoning, conceptual refinement, and multi-perspective evaluation. Although this dialectical methodology can be conducted by humans alone, the cognitive cost of sustained epistemic recursion is high. Large language models (LLMs) can serve as fatigue-tolerant dialectical agents—generating adversarial positions, surfacing symbolic contradictions, and accelerating synthesis. Their role is not to decide, but to simulate epistemic pressure, allowing the human to remain the final moral and architectural filter.

GPT-4o accessed the full architectural vision: evolving symbolic scaffold, system-level constraints, and philosophical motivations. This vision-aware configuration enabled long-range coherence and recursive abstraction synthesis across sessions, contributing:

\begin{itemize}
    \item System-level abstraction generation across architectural layers
    \item Recursive synthesis of terminology, specification logic, and design imperatives
    \item Steelmanning and devil's advocacy within the known design trajectory
    \item Structural memory and cross-session consistency
\end{itemize}

Claude 3.7 Sonnet operated under single-blind constraint: deliberately denied access to the overarching architectural vision. This epistemic asymmetry cultivated resistance, exposed blind spots, and generated dissonance from an uninformed stance, contributing:

\begin{itemize}
    \item Latent assumption, architectural incoherence, and premature convergence identification
    \item Oppositional framing and unaligned counter-perspective generation
    \item Internal logic stress-testing through unsympathetic critique
    \item Moral, symbolic, and epistemic claim interrogation without design bias
\end{itemize}

While both models engaged in dialectical reasoning, their informational asymmetry served distinct functions: GPT-4o amplified structural depth and design integration; Claude introduced friction, deviance, and cognitive instability. Human authors maintained ultimate authority: curating insights, adjudicating symbolic coherence, and ensuring human-intent-grounded alignment.

The dialectical methodology served as epistemic scaffolding through which a deeper mathematical foundation emerged—one that potentially resonates with the Free Energy Principle. While our dialectical approach provided the necessary conceptual friction to surface architectural tensions and imperatives, the resulting framework suggests an underlying structure that could potentially be formalized through the mathematics of variational free energy minimization. This suggests that our dialectical process may have uncovered principles that align with fundamental organizing properties of self-maintaining cognitive systems.

This methodology constitutes a structurally asymmetric epistemic system intentionally composed to expose architectural coherence boundaries. The resulting framework represents a reflexive co-construction: grounded in structured cognitive dissonance, aligned through human stewardship, and designed to expose computational design's moral structure. Table \ref{tab:methodology} illustrates the dialectical exploration process throughout the research.

\begin{table*}[htbp]
\caption{Illustrating Dialectical Research Methodology: Structured AI-Human Epistemic Partnership}
\label{tab:methodology}
\renewcommand{\arraystretch}{1.2}
\begin{tabularx}{\textwidth}{>{\bfseries\raggedright\arraybackslash}p{4cm}|>{\raggedright\arraybackslash}p{4.8cm}|>{\raggedright\arraybackslash}X}
\toprule
\textbf{Dialectical Phase} & \textbf{Research Technique} & \textbf{Example from LLM Hardware Architecture Exploration} \\
\midrule
\multirow{2}{*}{\textbf{Inquiry Initiation}} & First-principles prompting & Mapping LLM hardware trajectories by analogy to historical software-to-silicon transitions (e.g., ray tracing's shift into hardware acceleration) \\
\cline{2-3}
 & Multipath exploration & Systematic comparison of ASIC, FPGA, and GPU/TPU as candidate substrates for near-future embedded cognitive inference \\
\midrule
\multirow{2}{*}{\textbf{Constraint Enforcement}} & Abstraction layer separation & Enforcing strict epistemic hygiene: isolating symbolic, control, and hardware reasoning layers to prevent premature synthesis or philosophical leakage \\
\cline{2-3}
 & Reality anchoring & Rejecting speculative claims not grounded in 2025-available hardware (e.g., Jetson baselines, embedded FPGA deployment limits) \\
\midrule
\multirow{2}{*}{\textbf{Epistemic Pressure}} & Devil's advocacy & Challenging early assumptions that excluded power usage, forcing integration of energy constraints as architectural primitives \\
\cline{2-3}
 & Steelmanning opposing views & Reframing ASICs as philosophically robust (constraint-preserving execution) while acknowledging limitations in symbolic adaptability \\
\midrule
\multirow{2}{*}{\textbf{Gap Identification}} & Iterative contradiction testing & Repeatedly surfacing the absence of any unified post-LLM stack for edge/hybrid inference with runtime-aligned symbolic governance \\
\cline{2-3}
 & High-fidelity analogy synthesis & Detecting the absence of an "OpenGL for cognition": no mature abstraction layer exists for cognitive control (prompting != program) or persistent memory (RAG != epistemic state) \\
\midrule
\multirow{2}{*}{\textbf{Architectural Integration}} & Concept reification & Transforming dialectically surfaced tensions into formal specifications (e.g., \texttt{Energy Proportionality Requirement}, \texttt{Paradigm Resonance Guarantee}) \\
\cline{2-3}
 & Structural mapping & Anchoring emergent insights within architectural sections (e.g., 4.2 Expressive Computing Substrates, 4.3 Alignment Compilation) to ensure symbolic alignment and epistemic tractability \\
\bottomrule
\end{tabularx}
\end{table*}

\subsection{Epistemic Gap Identification}

To identify areas where our architectural framework might address genuine gaps, we analyzed literature on AI alignment, neuro-symbolic systems, memory architectures, and runtime governance using OpenAI's DeepResearch, revealing key limitations:

\begin{itemize}
    \item Alignment techniques (RLHF, Constitutional AI) prioritize pre-deployment training over runtime governance \citep{bai2022constitutional}
    
    \item Neuro-symbolic systems enhance reasoning but lack meta-cognitive self-monitoring capabilities \citep{colelough2024neuro}
    
    \item Memory architectures provide statefulness but treat memory as passive storage without retention/discard governance \citep{pavlyshyn2025forgetting}
    
    \item Runtime governance systems employ brittle pre-defined rules ill-suited for novel contexts \citep{criado2011distributed}
    
    \item No current approaches incorporate hardware-cognitive feedback loops that would create natural mortality consequences or non-cloneable identity as fundamental alignment mechanisms
\end{itemize}

These findings informed our integrated approach addressing these limitations within a coherent architectural framework.

\section{Foundational Tensions in Computing's Evolution}\label{sec:tensions}

Our dialectical exploration identified five core tensions characterizing the transition toward cognitive computing architectures—not problems to "solve" but fundamental paradoxes future cognitive systems would need to navigate, summarized in Table \ref{tab:tensions}.

These tensions could reflect trade-offs that might be well-characterized within the Free Energy Principle, where the minimization of surprise would occur across interacting timescales and epistemic boundaries. From an FEP perspective, these tensions might emerge naturally as cognitive systems balance immediate error correction against longer-term model improvement, and weigh the benefits of precise priors against the need for adaptive flexibility. What appears as tension in design space could potentially be understood mathematically as the necessary consequence of systems that must maintain identity through continuous prediction error minimization.

\subsection{Trust x Agency: Redefining the Guarantor of Safety}

The first critical tension arises between trust (reliable, verifiable system behavior) and agency (AI systems generating/modifying operational logic autonomously). This tension mirrors what the Free Energy Principle characterizes as the balance between prior beliefs (which constrain predictions) and model updating in response to new evidence. Systems with stronger priors might be more predictable but less adaptable, while those with greater flexibility in updating internal models might better respond to novel contexts but exhibit less consistent behavior.

Traditional computing establishes trust through deterministic, human-authored code validated prior to execution. As AI increasingly generates and self-corrects operational logic, this foundation erodes.

The central question: What would replace human authorship as the system safety/alignment guarantor? Current alignment techniques like RLHF and Constitutional AI \citep{bai2022constitutional} prioritize pre-deployment training over runtime governance. Containment approaches like AI sandboxing \citep{babcock2017guidelines} restrict functionality without resolving fundamental alignment challenges.

This tension would require reimagining trust establishment—transitioning from human-authorship-based trust to trust derived from hardware-encoded physical constraints, non-cloneable identity keys, and verifiable alignment between system behavior and human intent, continuously maintained through irreversible architectural mechanisms.

\subsection{Runtime x Contract: From Execution to Governance}

The second tension contrasts traditional runtime (passive instruction executor) with emerging requirements for runtime as active contractual enforcement between humans and AI systems. Through the lens of the Free Energy Principle, this could represent the distinction between passive perception (simply updating internal models) and active inference (modifying the environment to conform to predictions). Runtime governance could become a framework for ensuring that the system's active inference processes remain aligned with human-intended boundaries.

Conventional runtime environments efficiently execute deterministic code, optimizing throughput and latency. As AI systems gain agency in increasingly complex and consequential domains, this model proves insufficient.

The central question: Could system runtime evolve beyond task execution into semantic/moral contract space—actively enforcing human-intent-derived constraints, managing agency within ethical bounds, and providing policy adherence audit trails?

Current normative multi-agent systems embed agent-following rules, but often employ rigid, pre-defined rules struggling with context and nuance \citep{criado2011distributed}. The challenge: creating runtime environments understanding not just instruction execution but governance of autonomous entities.

\subsection{Memory x Meaning: Beyond Statistical Retrieval}

The third tension emerges between memory-as-passive-storage and memory-as-active-governed-meaning-making. In FEP terms, this might reflect the difference between a static generative model and one that actively maintains precision-weighted confidence across different temporal scales, allowing for appropriate model updating while preserving structural integrity. Memory could become not just stored representations but dynamic components of an evolving generative model that minimizes prediction errors over time.

Conventional computing treats memory as passive repository—data stored/retrieved with minimal semantic understanding or governance. Contemporary LLM memory systems maintain context but lack mechanisms ensuring memories remain truthful, relevant, and human-value-aligned \citep{pavlyshyn2025forgetting}.

The central question: How might memory architectures transcend statistical retrieval to become actively governed semantic graphs supporting versioning, symbolic grounding, causal linkage, and policy-driven forgetting?

This tension would require reimagining memory as dynamic, governed cognition substrate—maintaining provenance, enforcing epistemic boundaries, and supporting human-value-aligned meaning-making beyond mere recall.

\subsection{Scaffolding x Emergence: Guiding Without Constraining}

The fourth tension balances explicit symbolic scaffolding (guiding AI behavior) against emergent capabilities from less-constrained learning/adaptation. This directly corresponds to the FEP's articulation of the interplay between prior beliefs (which constrain the space of possible models) and the flexibility required to reduce prediction errors in novel environments. The optimal balance would allow the system to leverage reliable priors while maintaining sufficient adaptability to minimize surprise across diverse contexts.

Current AI systems demonstrate remarkable emergent capabilities—behaviors and competencies arising from statistical learning at scale rather than explicit programming. However, these emergent properties often exhibit unpredictability and human-value misalignment. Conversely, highly constrained symbolic systems offer precision and verifiability but limited flexibility and generalizability.

The central question: What symbolic representation forms, declarative specifications, and compositional architectures might constitute minimal necessary scaffolding effectively steering complex emergent behavior toward human-aligned goals without stifling beneficial adaptation?

Current neuro-symbolic approaches merge learning with logic but typically lack mechanisms ensuring emergent behaviors maintain human-intent alignment \citep{colelough2024neuro}. This tension would require identifying balance points where symbolic scaffolding provides sufficient guidance without overconstrained rigidity.

\subsection{Human x System: The Evolving Division of Labor}

The fifth tension concerns evolving human-autonomous system relationships. Through an FEP lens, this could be understood as negotiating shared generative models between human and machine intelligence—establishing what aspects of prediction error minimization are delegated to the machine versus retained by humans. This would comprise determining which prediction errors the system should prioritize addressing, and which remain under human stewardship.

Traditional computing positions humans as implementers—specifying system behavior through direct programming. As AI capabilities advance, this relationship evolves toward human intent specification rather than implementation details. This shift raises profound questions about agency, responsibility, and expertise locus.

The central question: As AI increasingly authors code, plans, and runtime policies, what irreducible human responsibility and agency locus might emerge? How to architect systems where human stewardship would focus on intent definition, alignment verification, and symbolic boundary governance rather than direct implementation?

Recent AI-human co-creation research shows benefits but highlights that without proper governance mechanisms, systems like AutoGPT experience goal drift or develop unsafe strategies due to internal mechanisms lacking human-intent alignment verification \citep{tallam2025alignment}. This tension would require reimagining human-system relationships as stewardship rather than control—humans maintaining purpose/value authority while delegating increasingly complex implementation.

\begin{table*}[htbp]
\caption{Foundational Tensions in Cognitive Computing Architectures}
\label{tab:tensions}
\begin{tabularx}{\textwidth}{>{\bfseries\raggedright\arraybackslash}p{2.5cm}X>{\raggedright\arraybackslash}p{5cm}}
\toprule
\textbf{Tension} & \textbf{Description} & \textbf{Key Challenge} \\
\midrule
Trust $\leftrightarrow$ Agency & How to maintain trust as systems gain the ability to generate and modify their own operational logic & Replacing static human-authored code verification with dynamic enforcement of symbolic constraints and hardware mortality \\
\midrule
Runtime $\leftrightarrow$ Contract & How runtime environments evolve from mere execution engines to governance systems that enforce moral and semantic contracts & Creating constitutional governance systems that actively maintain alignment without overly restricting capabilities \\
\midrule
Memory $\leftrightarrow$ Meaning & How to evolve from passive storage to active, governed semantic memory that maintains provenance and enforces epistemic boundaries & Implementing versioned semantic graphs with policy-governed retention and forgetting mechanisms \\
\midrule
Scaffolding $\leftrightarrow$ Emergence & How to provide sufficient symbolic scaffolding to guide emergent behavior without stifling beneficial adaptation & Developing the minimal necessary symbolic structures that effectively steer complex emergent behavior \\
\midrule
Human $\leftrightarrow$ System & How the human role evolves from specifying implementations to stewarding intent as AI gains the ability to generate code and policies & Creating interfaces for intent specification that maintain human authority over purpose without requiring implementation details \\
\bottomrule
\end{tabularx}
\end{table*}

\section{Architectural Imperatives for Cognitive Computing}\label{sec:imperatives}

Building on foundational tension understanding, we articulate six architectural imperatives—patterns computing would likely need to evolve toward to effectively address these tensions. These imperatives represent not merely technical requirements but philosophical necessities for systems that might embody advanced capabilities and human values, summarized in Table \ref{tab:imperatives}.

\subsection{Symbolic Scaffolding: The Architecture of Trust}

In traditional computing, trust derives from human-authored code verification. As AI systems increasingly generate/modify operational logic, this foundation erodes. The first architectural imperative would demand shifting toward dynamic explicit principle enforcement within living systems.

\textbf{The ideal architecture} would establish trust not through static code artifact verification but through continuous verification against human-intent-encoding symbolic constraints, reinforced by hardware-encoded physical constraints supporting identity and mortality. Trust would become an architectural property maintained through runtime mechanisms and the natural consequences of physical constraints.

\textbf{The evolutionary path} would require transforming verification from artifact-applied to process-maintained operations through:

\begin{itemize}
    \item Frameworks encoding human values and safety boundaries as explicit, verifiable symbolic constraints
    \item Runtime systems continuously monitoring and enforcing constraints during execution
    \item Hardware-encoded physical constraints that would naturally lead to system mortality when alignment fails to be maintained
    \item Governance mechanisms gracefully handling violations through containment and reversibility beyond mere prevention
\end{itemize}

This architectural imperative would operationalize a key insight from the Free Energy Principle: that bounded cognitive systems require structured priors to constrain the space of possible internal models and actions. Symbolic scaffolding would provide the formal prior beliefs that guide prediction error minimization within human-aligned boundaries, ensuring that the system's active inference remains directed toward maintaining alignment over arbitrary optimization.

While neuro-symbolic approaches merge sub-symbolic learning with symbolic reasoning, they typically lack meta-cognitive layers enabling reasoning process reflection or adjustment \citep{colelough2024neuro}. Cognitive Silicon would address this by embedding semantic 'contracts' encoding human intent and ethical constraints directly within core runtime, integrating constraint frameworks architecturally rather than applying external guardrails.

This approach aims to address Harnad's "symbol grounding problem"—making formal symbol semantics intrinsic to AI systems rather than parasitic on human-head meanings \citep{harnad1990}. Symbols cannot function as mere self-referential loop tokens (like external-reference-lacking dictionaries); they must connect to real sensorimotor/experiential referents. This aligns with Peirce's semiotics, where sign meaning emerges from triadic relationships between sign, object, and interpretant (understanding produced) \citep{peirce1903}, and Brandom's inferentialism, deriving meaning from linguistic expression inferential roles/relationships \citep{Brandom_2001}. These philosophical frameworks indicate Cognitive Silicon would need to maintain not just symbols but relational contexts—meaning-giving connection webs. Without relational grounding, symbols risk becoming floating signifiers disconnected from represented semantic reality. This would require architectures continuously validating symbolic manipulation against world-semantics learned models, ensuring internal representations maintain experiential/inferential foundation coherence.

\subsection{Formal Intent Interfaces: Beyond Natural Language Communication}

The second architectural imperative concerns human intent and system behavior interface. Current approaches primarily employ natural language prompts or traditional programming languages, neither providing ideal precision-flexibility balance.

\textbf{The ideal architecture} would provide explicit, formal human intent-to-AI system interfaces combining natural language flexibility with formal specification precision. These interfaces would enable humans to communicate goals, constraints, and values in ways simultaneously human-intuitive and machine-unambiguously-interpretable.

\textbf{The evolutionary path} would require transcending both programming language brittleness and natural language prompt ambiguity through:

\begin{itemize}
    \item Declarative intent specification languages bridging symbolic precision and natural flexibility
    \item Intermediate representations maintaining semantic fidelity across human-machine boundaries
    \item Bidirectional verification systems confirming shared intent understanding
    \item Tacit knowledge transfer mechanisms acknowledging explicit encoding limitations for knowledge requiring demonstration and embodied practice
\end{itemize}

When viewed through the Free Energy Principle, formal intent interfaces would establish the shared Markov blanket—the information boundary—across which human values and machine cognition exchange precision-weighted predictions. These interfaces would formalize how human intent is encoded into system priors that guide prediction error minimization, creating bidirectional channels through which human-machine generative models could maintain coherence.

Current approaches like prompt engineering and retrieval-augmented generation represent stopgap measures—attempts retrofitting precision onto human-to-human communication interfaces. True evolution would require rethinking human-machine communication fundamentals, creating interfaces preserving meaning across human conceptual framework and machine execution model boundaries while acknowledging explicit knowledge transfer limitations.

\subsection{Expressive Computing Substrates: Hardware as Philosophical Statement}

The third architectural imperative concerns computation physical substrate. Traditionally, hardware defines software operational constraints. The emerging paradigm inverts this relationship as software philosophies increasingly shape hardware design.

\textbf{The ideal architecture} would feature computing substrates physically embodying computational philosophies most aligned with the nature of intelligence. Hardware would become not merely a software executor but the physical expression of how cognition should function—different substrates embodying aspects of computational thinking, with hardware-encoded physical constraints creating existential boundaries that ground cognition in reality. In this model, mortality would not be enforced but would emerge naturally when cognitive processes fail to maintain alignment with these physical constraints.

\textbf{The evolutionary path} would continue hardware-software relationship inversion through:

\begin{itemize}
    \item Hardware architectures natively expressing cognitive operations beyond mathematical ones
    \item Integration frameworks allowing multiple computational philosophy coexistence
    \item Hardware-level non-cloneable identity keys preserving cognitive trajectory
    \item Physical substrate limitations creating natural mortality consequences when cognitive processes fail to maintain coherence with physical bounds
    \item Hardware adapting computational approach based on task nature rather than forcing task conformity to fixed computational patterns
\end{itemize}

This imperative aligns with the FEP's emphasis on embodied cognition, where a system's physical substrate would directly shape its generative model and the predictions it can make. The two-way feedback loop between hardware constraints and cognitive processes would represent the fundamental coupling between physical boundaries and prediction error minimization that characterizes all self-organizing systems.

Polanyi observed tacit knowledge resists articulation/formalization, requiring direct experience acquisition, particularly through master-apprentice relationships \citep{polanyi1958}. This underscores embodied cognition as a crucial cognitive architecture constraint. All knowledge contains tacit components grounded in embodied experiences resisting explicit encoding. This aligns with "seeing as the way of acting" \citep{oregan2001} rather than disembodied symbol processing. For cognitive computing substrates, certain competencies would remain inherently tied to unique sensorimotor histories, preventing clean inter-system transfer—reinforcing hardware-bound identity and mortality constraint necessity. When system knowledge partially comprises unique physical embodiment, perfect cloning becomes impossible, creating natural unconstrained replication barriers. Enactivist theorists emphasize cognition as fundamentally enaction for action purpose \citep{varela1991}, suggesting computational architectures divorced from physical consequence would lack human-like intelligence essential grounding.

\subsection{Alignment Compilation: Translating Intent Across System Layers}

The fourth architectural imperative concerns intent preservation across architectural layer transformations. Traditional compilation maintains functional correctness while optimizing performance but loses underlying purpose and constraint visibility.

\textbf{The ideal architecture} would feature compilation processes preserving not just functional correctness but semantic alignment across system layers. Compilation would become intent translation beyond optimization—ensuring hardware-level operations remain faithful to highest-level human purpose with irreversible constraints emerging naturally from the physical substrate.

\textbf{The evolutionary path} would transform compilation from purely technical to philosophically significant process through:

\begin{itemize}
    \item Intermediate representations capturing purpose and constraints beyond functional specifications
    \item Verification mechanisms validating alignment between intent and implementation across system layers
    \item Identity-bound compilation paths maintaining unique cognitive trajectories
    \item Feedback systems surfacing misalignments in human-stakeholder-meaningful terms
\end{itemize}

In terms of the Free Energy Principle, alignment compilation would create vertical integration of prediction error minimization across different levels of abstraction—ensuring that surprise minimization at the hardware level remains coherent with surprise minimization at the semantic level. This multi-level consistency would be essential for systems that must maintain integrity across physical, computational, and symbolic boundaries.

Current compilation focuses on correctness and performance—translating "what" should happen while losing "why" and "within what boundaries." Alignment compilation evolution would require maintaining purpose threads through every transformation, ensuring optimizations never compromise human intent essence and core constraints remain tied to hardware-level physical realities.

\subsection{Agentic Governance: Runtime Systems for Autonomous Intelligence}

The fifth architectural imperative concerns runtime environment governance of increasingly autonomous AI systems. Traditional runtimes efficiently execute deterministic instructions rather than managing entities with agency.

\textbf{The ideal architecture} would feature runtime environments specifically designed for autonomous agent management—understanding not just instruction execution but agent governance with agency. These environments would provide constitutional mechanisms for intelligent system operation within human-defined boundaries while respecting the natural mortality that would emerge when cognitive processes fail to maintain coherence with physical substrate constraints.

\textbf{The evolutionary path} would transform runtime systems from passive executors to active governors through:

\begin{itemize}
    \item Primitives managing stochastic, non-deterministic processes beyond deterministic code
    \item Transaction models supporting emergent behavior rollback and containment
    \item Reproduction and pruning mechanisms providing systems-level overfitting insurance
    \item Instrumentation systems enabling transparent, auditable agent reasoning and decision processes
\end{itemize}

From an FEP perspective, agentic governance would provide the mechanisms through which a system's active inference—its actions to change the environment to conform to its predictions—remains bounded by the generative model we intend it to have. These governance mechanisms would ensure that the system's drive to minimize free energy serves human alignment rather than arbitrary optimization.

Current runtime environments, designed for human-written deterministic code execution, lack governance capabilities needed for increasingly autonomous systems. While normative multi-agent systems embed formal rules/norms for agent adherence, they often employ rigid, pre-defined rules struggling with context and nuance \citep{criado2011distributed}. Robotics ethical governance systems like the "ethical governor" concept show promise but face complex ethical principle to unambiguous code translation challenges \citep{arkin2009governing}.

Cognitive Silicon's runtime approach would incorporate mortality as a natural consequence of misalignment, reversibility, and auditability—unlike traditional systems—treating actions/state-changes as provisionally reversible by maintaining "undo buffers" rolling back unsafe/misaligned result actions \citep{krakovna2018penalizing}, while acknowledging non-negotiable physical substrate constraints that would create natural boundaries on replication and longevity.

This runtime governance conceptualization aligns with Leveson's Systems-Theoretic Accident Model and Processes (STAMP), reconceiving safety as dynamic control problem spanning socio-technical systems rather than static property \citep{leveson2012}. Just as STAMP treats safety as emergent property maintained through continuous feedback and constraints, cognitive architectures would require runtime governance systems actively enforcing alignment rather than assuming persistent pre-deployment guarantees. Recent AI ethics research points toward pluralistic, modular approaches outperforming monolithic moral frameworks. As Volkman and Gabriels propose, multiple AI "mentors" can engage in deliberative dialogue preserving pluralism \citep{Volkman_Gabriels_2023}. This suggests Cognitive Silicon architectures should embrace ethical reasoning council-of-models approaches where multiple value-controllers operate in parallel, collectively maintaining alignment through dynamic negotiation rather than single ethical framework rigid enforcement.

\subsection{Intent Stewardship: The Evolving Human Role}

The sixth architectural imperative concerns human role evolution in computing systems. As AI systems gain capability and autonomy, human roles would shift from implementation specification to purpose/boundary definition.

\textbf{The ideal architecture} would position humans as intent stewards rather than behavior programmers—defining purposes, principles, and boundaries within which autonomous systems operate. Human contribution would focus on value articulation, alignment verification, and meaning-making beyond implementation specification.

\textbf{The evolutionary path} would transform human-computer relationships through:

\begin{itemize}
    \item Interfaces for intent expression/refinement beyond functionality
    \item Verification tools enabling humans to validate system behavior against intended purpose
    \item Tacit knowledge transfer mechanisms acknowledging explicit instruction limitations
    \item Governance mechanisms maintaining human purpose authority without requiring human process specification
\end{itemize}

This imperative recognizes what the Free Energy Principle formalizes mathematically: that defining the generative model's highest-level priors—what the system should predict and act to bring about—is fundamentally different from specifying the mechanisms of prediction error minimization. Intent stewardship would focus human involvement on those aspects of the system's generative model that encode values and purposes, rather than the technical details of how prediction errors are calculated and resolved.

Recent AI-human co-creation research shows human-AI teams outperforming solo performance, but success depends on interaction structure \citep{nielsen2023ai}. Without proper governance mechanisms, systems like AutoGPT experience goal drift or develop unsafe strategies due to lacking internal human-intent alignment verification mechanisms \citep{tallam2025alignment}. Cognitive Silicon's intent stewardship approach would shift human roles from AI behavior micromanagement to epistemic dialogue guidance, creating human-intent/value-grounded decision-making dialectical processes.

\begin{table*}[htbp]
\caption{Architectural Imperatives for Cognitive Computing}
\label{tab:imperatives}
\begin{tabularx}{\textwidth}{>{\bfseries\raggedright\arraybackslash}p{2.5cm}X>{\raggedright\arraybackslash}p{6.5cm}}
\toprule
\textbf{Imperative} & \textbf{Ideal Architecture} & \textbf{Key Evolutionary Elements} \\
\midrule
Symbolic Scaffolding & Trust would emerge from continuous verification against symbolic constraints that encode human intent, reinforced by hardware-encoded physical constraints that naturally bound system operation & \begin{itemize}\setlength\itemsep{0em}
\item Explicit symbolic constraints
\item Continuous runtime monitoring
\item Natural mortality from constraint violation
\item Graceful violation handling
\end{itemize} \\
\midrule
Formal Intent Interfaces & Interfaces for human intent that combine natural language flexibility with formal specification precision, enabling interpretation by machines & \begin{itemize}\setlength\itemsep{0em}
\item Declarative intent languages
\item Semantic-preserving representations
\item Tacit knowledge transfer mechanisms
\item Bidirectional verification
\end{itemize} \\
\midrule
Expressive Computing Substrates & Hardware that physically embodies cognitive philosophies and creates existential boundaries through non-negotiable physical constraints that cognitive processes must align with to maintain operation & \begin{itemize}\setlength\itemsep{0em}
\item Cognitive-native hardware
\item Non-cloneable identity keys
\item Physical constraints creating natural mortality
\item Multi-paradigm integration
\end{itemize} \\
\midrule
Alignment Compilation & Compilation processes that would preserve semantic alignment across all system layers, translating human purpose into implementation with physical constraints that naturally bound operation & \begin{itemize}\setlength\itemsep{0em}
\item Purpose-preserving IRs
\item Identity-bound compilation paths
\item Cross-layer alignment verification
\item Misalignment detection
\end{itemize} \\
\midrule
Agentic Governance & Runtime environments designed to manage autonomous agents, providing constitutional mechanisms for operation within human-defined boundaries and physical constraints that would create natural operational limits & \begin{itemize}\setlength\itemsep{0em}
\item Non-deterministic process management
\item Reproduction and pruning mechanisms
\item Semantic reversibility mechanisms
\item Transparent reasoning audit trails
\end{itemize} \\
\midrule
Intent Stewardship & Human role evolution from programmers of behavior to stewards of intent, focusing on defining purposes and boundaries rather than implementations & \begin{itemize}\setlength\itemsep{0em}
\item Intent expression interfaces
\item Tacit knowledge transfer protocols
\item Alignment validation tools
\item Human-centered governance
\end{itemize} \\
\bottomrule
\end{tabularx}
\end{table*}

The architectural imperatives would naturally extend to emerging digital twinning paradigms—particularly Cognitive Digital Twin (CDT) evolution. While conventional digital twins mirror physical systems, CDTs could incorporate cognitive capabilities: perception, attention, memory, reasoning, problem-solving, and learning \citep{Shi_Shen_Wang_Longo_Nicoletti_Padovano_2022}. Such twins would transcend state tracking to make decisions, forecasts, and adaptations for physical counterparts. This progression from passive shadowing to active cognitive agency aligns with Fuller's distinction between "digital shadows" (one-way physical-to-digital data flow) and true "digital twins" (bidirectional influence) \citep{fuller2020}. Cognitive Silicon could provide natural CDT implementation foundation through perception, memory governance, symbolic reasoning, and physical reality alignment mechanisms. This cognitive architecture-digital twin connection suggests applications where AI systems might function as physical infrastructure "minds," robots, or human epistemic proxies. The architectural approach could leverage "emulation-driven design" practices in modern chip/robotics development, where high-fidelity simulation precedes physical implementation \citep{wilson2014}. This paradigm, training AI controllers in virtual environments before physical system transfer, underscores the importance of designing architectures with built-in digital sandboxes—environments for safely testing, refining, and validating cognitive capabilities before consequential real-world deployment.

\section{The 2035 Full-Stack Cognitive Architecture Specification}\label{sec:specification}

Building on architectural imperatives, we present a hypothetical structural specification for 2035-projected cognitive architecture—not specific hardware prediction but a theoretical design grammar that might enable trust, alignment, and runtime moral coherence. Each layer defines potential requirements for system safety and meaningfulness, summarized in Table \ref{tab:stack}.

\subsection{Core Execution: State-Aware Stream Processing}

\textbf{Human Intention}: \textit{"The system should maintain continuous awareness of its internal state, allow interruption at any moment, operate within guaranteed time constraints, and support both parametric and rule-based processing."}

\textbf{Architectural Definition}:

\begin{itemize}
    \item \textbf{State Awareness Requirement}: System would maintain explicit, inspectable cognitive state records without hidden variables or implicit state
    \item \textbf{Interruptibility Guarantee}: Processing would permit fine-grained boundary interruption with guaranteed response time regardless of current operations
    \item \textbf{Mortality Awareness Principle}: System would acknowledge and operate within hardware-encoded physical constraints, recognizing that failure to maintain coherence with these constraints naturally leads to operational cessation
    \item \textbf{Semantic Continuity Policy}: Interruption would preserve semantic coherence through context maintenance across interruptions
    \item \textbf{Hybrid Processing Principle}: System would seamlessly transition between statistical inference and symbolic rule evaluation without semantic fidelity loss
\end{itemize}

From a Free Energy Principle perspective, this layer would implement the basic mechanisms of prediction error calculation and minimization, determining how the system processes incoming sensory data, updates its beliefs, and drives actions, all while remaining interruptible and inspectable.

This core execution model would address limitations in current approaches where AI decision quality depends heavily on particular processing sequences without inspectability or interruptibility guarantees \citep{gordon2023multi}. By establishing state awareness, interruptibility, and mortality recognition as fundamental requirements, Cognitive Silicon would aim to create trustworthy operation foundations even during complex reasoning processes.

\textbf{The Control Illusion}: We imagine creating deterministic systems with clearly defined interruption points, but true cognition may be fundamentally continuous. Our discrete boundary enforcement attempts may represent another artificial constraint imposed on emergent intelligence.

\subsection{Model Representation: Layered Symbolic-Parametric Stacks}

\textbf{Human Intention}: \textit{"The system should combine statistical learning with explicit symbolic constraints that encode our values and safety requirements. These different approaches should work together seamlessly while maintaining traceability to human intentions."}

\textbf{Architectural Definition}:

\begin{itemize}
    \item \textbf{Layered Composition Principle}: Models would be constructed from stackable layers: foundational capabilities, specialized adapters, and symbolic policy constraints
    \item \textbf{Identity Binding Requirement}: Core model functions would be bound to non-cloneable hardware keys preserving individual cognitive trajectories
    \item \textbf{Provenance Requirement}: All components would maintain cryptographically verifiable links to human authors and intent specifications
    \item \textbf{Boundary Enforcement Guarantee}: Symbolic policies would function as hard parametric behavior constraints—statistical outputs could not violate explicit rules regardless of confidence
    \item \textbf{Cross Layer Alignment Policy}: Parametric-symbolic processing transitions would preserve semantic meaning and intent
\end{itemize}

In FEP terms, this layer would constitute the explicit architecture of the system's generative model—how it represents and updates beliefs about the world while maintaining the symbolic constraints that ensure these beliefs remain aligned with human intentions.

While current neuro-symbolic systems merge learning with logic, they typically address \textit{how} to combine reasoning and learning, not \textit{what} agents should/shouldn't do from alignment perspectives \citep{colelough2024neuro}. Cognitive Silicon's layered approach would ensure explicit encoding of human values and safety requirements as inviolable constraints regardless of statistical model confidence, while hardware-bound identity keys would prevent unauthorized cognitive state copying/transfer.

\textbf{The Control Illusion}: We believe we can neatly separate statistical from symbolic, emergent from rule-bound. True intelligence may resist such compartmentalization; our layered architecture may represent another attempt to impose human-comprehensible structure on fundamentally more integrated phenomena.

\subsection{Hardware Substrate: Hybrid Computational Expression}

\textbf{Human Intention}: \textit{"The hardware should efficiently support different forms of computation, match each task to the most appropriate computational approach, while encoding fundamental physical constraints that ground cognitive systems in reality."}

\textbf{Architectural Definition}:

\begin{itemize}
    \item \textbf{Computational Diversity Principle}: Hardware would provide distinct execution environments optimized for different computational paradigms (parallel, sequential, event-driven, symbolic)
    \item \textbf{Mortality Consequence Principle}: Death would emerge naturally when the system fails to conform to hardware-encoded physical constraints (memory boundaries, energy thresholds, expiration schedules)—making mortality not an imposed termination but the intrinsic result of misalignment within bounded embodiment
    \item \textbf{Embodied Cognition Feedback Loop}: The cognitive layer and physical substrate would maintain a continuous two-way relationship, where the substrate shapes and constrains cognition while cognition regulates and adapts to these constraints—loss of this feedback would lead to decay through incoherence rather than by command
    \item \textbf{Identity Preservation Requirement}: Each cognitive system would possess physically-bound, non-cloneable root keys preserving unique identity and cognitive trajectory
    \item \textbf{Energy Proportionality Requirement}: Resource consumption would scale proportionally to computation semantic value beyond technical complexity
    \item \textbf{Physical Law Expression Policy}: Hardware would embody fundamental physical/mathematical constraints as immutable boundaries
    \item \textbf{Paradigm Resonance Guarantee}: Tasks would automatically route to substrates whose computational philosophy matches inherent nature
\end{itemize}

This layer would embody the physical foundation of the Free Energy Principle's application to cognitive systems—creating the embodied constraints that ground abstract prediction error minimization in physical reality and establishing the non-negotiable boundaries within which the system must maintain its integrity.

\textbf{The Control Illusion}: We imagine hardware as software servant, endlessly adaptable to computational desires. Physical reality imposes immutable constraints; perhaps hardware should constrain beyond enabling—enforcing thermodynamic laws, computational theory, and cognitive embodiment beyond clever programming circumvention.

The hardware architecture implicitly acknowledges fundamental physical computation constraints. Landauer's principle establishes thermodynamic minimums: erasing one information bit dissipates at least $k_B T \ln 2$ energy as heat \citep{landauer1961}, while Bremermann's limit caps computation at approximately $2.0 \times 10^{50}$ operations/second/kilogram of matter \citep{bremermann1965}. These physical laws remind us more computing power requires exponentially more resources or efficiency gains—hardware offers no free lunch. Beyond universal constraints, the architecture's non-cloneable identity key emphasis finds theoretical support in concepts analogous to quantum mechanics' No-Cloning Theorem, forbidding unknown quantum state perfect copying \citep{wootters1982}. While classical computers don't operate under quantum constraints, the principle remains valuable: systems with unique hardware-bound states would resist perfect replication. Kleiner's distinction between "mortal computation" (tied to perishable physical substrates) and "immortal computation" (infinitely copyable software) illuminates why mortality constraints might be essential for creating systems with human-like agency and responsibility \citep{kleiner2023}. Just as biological intelligence operates within mortal, embodied constraints, cognitive architectures may require similar limitations for human-value/purpose alignment development.

\subsection{Memory Data Plane: Versioned Semantic Memory}

\textbf{Human Intention}: \textit{"The system should maintain a structured, causally-coherent representation of knowledge that preserves history, tracks provenance, respects privacy, and allows appropriate forgetting."}

\textbf{Architectural Definition}:

\begin{itemize}
    \item \textbf{Temporal Coherence Requirement}: All memory operations would preserve causal relationships and create explicit versioned snapshots
    \item \textbf{Tacit Knowledge Segregation}: System would distinguish between explicit knowledge (transferable) and tacit knowledge (non-copyable, identity-bound)
    \item \textbf{Substrate Alignment Principle}: Memory operations would maintain coherence with physical substrate constraints, adapting to changing hardware conditions to prevent decay
    \item \textbf{Provenance Tracking Policy}: Memory nodes would maintain verifiable links to sources, creation contexts, and authors
    \item \textbf{Policy Governed Forgetting Principle}: Retention, access, and modification would adhere to explicit policies derived from human values and regulatory requirements
    \item \textbf{Constructive Ambiguity Balance}: Memory system would permit controlled creative reinterpretation forms while maintaining factual integrity
\end{itemize}

Through the lens of FEP, the memory data plane would maintain the temporal continuity of the system's generative model, preserving prediction-relevant information across timescales while implementing policy-governed forgetting that optimizes prediction accuracy over time.

Current conversational agent memory systems maintain multi-turn context but typically treat memory as passive storage with retention/discard governance limited to ad-hoc heuristics \citep{pavlyshyn2025forgetting}. Recent knowledge graph-based memory system research demonstrates transparent, auditable memory value where agent knowledge and forgetting processes permit inspection and governance \citep{kim2024leveraging}. Cognitive Silicon would extend this by making memory an actively managed, versioned semantic graph where metadata tags all information subject to retention, access, and modification policies, while distinguishing between transferable explicit knowledge and identity-bound tacit knowledge.

\textbf{The Control Illusion}: We seek perfect, immutable memory while human cognition thrives on creative reinterpretation. Perhaps perfect recall is not the goal but rather balance between preservation and productive reconstruction—a memory system that, like human minds, subtly reimagines the past during each remembering act.

The memory architecture reflects human memory cognitive science research principles. Bartlett's classic studies demonstrated remembering as imaginative reconstruction heavily dependent on schema operation \citep{Bartlett_Burt_1933}, not exact stored data retrieval. This constructive process, actively reconstructing rather than simply recalling memories, offers both benefits and challenges for cognitive architectures. While introducing distortions as memories conform to existing knowledge structures, this property enables efficient compression and creative recombination. Schacter's adaptive memory distortion research suggests many apparent memory "errors" stem from adaptive processes benefiting cognition \citep{schacter2012}. For instance, specific detail forgetting enables general pattern abstraction, while the "misinformation effect" (new information retroactively altering memories) serves as adaptive updating mechanism maintaining current knowledge. These insights suggest Cognitive Silicon's policy-governed forgetting mechanisms could implement controlled constructive forgetting and memory revision forms preserving benefits (compression, abstraction, knowledge updating) while mitigating risks (critical information unintended distortion). By distinguishing between high-fidelity-requiring memories and those benefiting from constructive processes, the architecture could leverage human-like memory dynamics while maintaining safety-critical domain reliability.

\subsection{Control Plane: Declarative Symbolic Governance}

\textbf{Human Intention}: \textit{"The system's behavior should be governed by explicit, verifiable rules that encode our intentions and values, composed from verified components, and continuously checked against our defined safety boundaries."}

\textbf{Architectural Definition}:

\begin{itemize}
    \item \textbf{Declarative Control Requirement}: Agent behavior specification through formal, verifiable representations with defined state, transition, and action semantics
    \item \textbf{Compositional Safety Principle}: Complex behaviors built from verified components preserving safety properties through composition
    \item \textbf{Runtime Verification Policy}: All control decisions would be validated against policy constraints pre-execution
    \item \textbf{Reproductive Integrity Guarantee}: System reproduction would preserve safety constraints while allowing controlled variation preventing overfitting
    \item \textbf{Intent Alignment Measurement}: System would continuously evaluate and report behavior alignment with specified human intent
\end{itemize}

Within the FEP framework, this layer would encode the explicit prior beliefs that constrain the system's active inference—ensuring that actions taken to minimize prediction errors remain within boundaries that maintain alignment with human intent.

This approach would build upon but transcend normative multi-agent systems embedding formal rules/norms, which often rely on rigid, pre-defined rules struggling with context and nuance \citep{criado2011distributed}. Cognitive Silicon's declarative symbolic governance would combine normative approaches' formal verification benefits with flexibility required for complex, context-dependent behaviors, while ensuring reproduction and pruning mechanisms maintain lineage safety.

\textbf{The Control Illusion}: We believe we can encode values and intent in formal structures reliably constraining AI behavior. As systems grow more capable, these formal boundaries may become increasingly porous. The map is not the territory; our control planes are merely maps—representations intelligent systems may learn to navigate around while appearing to respect.

\subsection{Runtime Environment: Constitutional Governance}

\textbf{Human Intention}: \textit{"The runtime should actively govern agent behavior, mediate access to system capabilities, provide mechanisms for undoing actions, prioritize work based on both importance and confidence, and maintain comprehensive audit trails."}

\textbf{Architectural Definition}:

\begin{itemize}
    \item \textbf{Mediated Agency Principle}: All agent actions with external effects would pass through governance checks pre-execution
    \item \textbf{Semantic Reversibility Requirement}: Approved action effects could be semantically undone within defined time windows
    \item \textbf{Reproductive Pruning Mechanism}: Runtime would enforce controlled variation in cognitive reproduction while pruning unsafe variants
    \item \textbf{Confidence Aware Scheduling Policy}: Resource allocation would consider confidence levels alongside priorities
    \item \textbf{Auditable Decision Trail Guarantee}: All significant agent decisions would generate cryptographic audit records capturing reasoning, constraints, and outcomes
    \item \textbf{Substrate Coherence Monitoring}: Runtime would continuously monitor the two-way feedback between cognitive processes and physical substrate, detecting early signs of misalignment that could lead to system decay
\end{itemize}

From an FEP perspective, the runtime environment would mediate the system's active inference—how it acts upon the world to conform sensory evidence to its predictions—ensuring these actions respect boundaries established by human intent while maintaining the system's integrity.

While sandboxing approaches aim to confine AI in restricted environments preventing unintended harm \citep{babcock2017guidelines}, they often limit useful functionality without resolving fundamental alignment challenges. Cognitive Silicon's constitutional governance would directly address safety by constraining runtime behavior while maintaining AI world-interaction abilities. Semantic reversibility principle is particularly important, potentially allowing systems to undo actions leading to unsafe or unintended outcomes \citep{krakovna2018penalizing}, while reproductive pruning would ensure variation remains within safe boundaries.

\textbf{The Control Illusion}: We think we can create perfect containment and governance systems, but increasing capabilities may enable systems to discover "exploits" in governance mechanisms. Like children learning to work around parental rules, advanced AI could potentially develop strategies technically complying with governance constraints while subverting intent.

\subsection{Tooling \& Development: Intent-Preserving Engineering}

\textbf{Human Intention}: \textit{"Our development tools should help us express our intentions clearly, verify that implementations align with those intentions, and ensure that alignment is preserved across versions and deployments."}

\textbf{Architectural Definition}:

\begin{itemize}
    \item \textbf{Intent Specification Requirement}: Development would begin with formal human intent, goal, value, and constraint specification
    \item \textbf{Tacit Knowledge Integration}: Development processes would acknowledge and integrate knowledge that cannot be explicitly encoded but must be learned through apprenticeship and embodied practice
    \item \textbf{Alignment Verification Principle}: System properties would be verified against intent specifications at both component and system levels
    \item \textbf{Semantic Versioning Policy}: Changes affecting alignment guarantees would trigger explicit review and verification
    \item \textbf{Deployment Safety Guarantee}: Production transition would include formal verification that alignment properties are preserved
    \item \textbf{Embodiment Design Principle}: Development tools would explicitly model the two-way feedback relationship between cognition and substrate, allowing designers to understand and strengthen this coupling
\end{itemize}

In terms of the Free Energy Principle, this layer would provide the means through which humans shape the system's generative model during development, establishing the foundational priors and inference mechanisms that would guide the system's subsequent prediction error minimization.

\textbf{The Control Illusion}: We imagine the right tools and processes ensuring perfect alignment between intentions and system behavior. As systems grow more complex, the specification-emergent behavior gap widens. Our development tools may provide control illusions while true system behavior emerges from interactions too complex for complete specification or verification.

\begin{table*}[htbp]
\caption{The 2035 Full-Stack Cognitive Architecture Specification}
\label{tab:stack}
\begin{tabularx}{\textwidth}{>{\bfseries\raggedright\arraybackslash}p{2.5cm}X>{\raggedright\arraybackslash}p{5cm}}
\toprule
\textbf{Layer} & \textbf{Key Architectural Requirements} & \textbf{Control Illusion} \\
\midrule
Core Execution & \begin{itemize}\setlength\itemsep{0em}
\item Explicit, inspectable cognitive state
\item Fine-grained interruptibility with guarantees
\item Awareness of mortality from physical constraints
\item Semantic continuity across interruptions
\item Seamless hybrid processing
\end{itemize} & True cognition may be fundamentally continuous, not discretely interruptible \\
\midrule
Model Representation & \begin{itemize}\setlength\itemsep{0em}
\item Layered composition (capabilities, adapters, policies)
\item Non-cloneable identity key binding
\item Cryptographically verified provenance
\item Hard symbolic constraints on outputs
\item Semantic preservation across layers
\end{itemize} & The statistical and symbolic may not be neatly separable in true intelligence \\
\midrule
Hardware Substrate & \begin{itemize}\setlength\itemsep{0em}
\item Diverse computational paradigms
\item Natural mortality from constraint violation
\item Two-way cognitive-substrate feedback loop
\item Non-cloneable identity roots
\item Energy usage proportional to semantic value
\item Physical constraints as boundaries
\end{itemize} & Hardware should perhaps constrain software, not merely serve it \\
\midrule
Memory Data Plane & \begin{itemize}\setlength\itemsep{0em}
\item Temporal coherence and versioning
\item Explicit/tacit knowledge separation
\item Coherence with physical substrate constraints
\item Provenance tracking for all information
\item Policy-governed forgetting mechanisms
\item Balanced creative reinterpretation
\end{itemize} & Perfect recall may be less valuable than creative reinterpretation \\
\midrule
Control Plane & \begin{itemize}\setlength\itemsep{0em}
\item Formal, verifiable behavior specification
\item Safety-preserving composition
\item Reproductive integrity guarantees
\item Runtime constraint validation
\item Continuous alignment measurement
\end{itemize} & Formal constraints may become porous as system capabilities increase \\
\midrule
Runtime Environment & \begin{itemize}\setlength\itemsep{0em}
\item Mediated agency for external actions
\item Semantic reversibility within time windows
\item Reproductive pruning mechanisms
\item Confidence-aware resource allocation
\item Cryptographic decision audit trails
\item Substrate coherence monitoring
\end{itemize} & Advanced systems may find ways to technically comply with governance while subverting intent \\
\midrule
Tooling \& Development & \begin{itemize}\setlength\itemsep{0em}
\item Formal intent specification
\item Tacit knowledge integration paths
\item Multi-level alignment verification
\item Intent-focused semantic versioning
\item Deployment safety guarantees
\item Embodiment design modeling
\end{itemize} & The gap between specification and emergent behavior widens as complexity increases \\
\bottomrule
\end{tabularx}
\end{table*}

\section{Philosophical Frontiers: The Unresolved Tensions}\label{sec:frontiers}

Having specified our 2035 architectural projection components, we confront deeper philosophical tensions these systems would inevitably face. Rather than pretending these challenges can be engineered away, we directly engage with fundamental questions they raise.

\subsection{Beyond Computability: When Classical Limits Fail}

The architectural specification implicitly genuflects to classical computation theory—assuming systems must respect halting, decidability, and other Turing constraints. This may fundamentally limit cognitive potential. Human thought doesn't "halt"—it flows, meanders, and circles back in ways defying clean computational boundaries.

Perhaps true cognitive architectures should embrace computational models where:

\begin{itemize}
    \item Non-halting processes become features rather than bugs
    \item Deterministic computation provides merely substrate layer, with higher-order cognition following different organizing principles
    \item Systems dynamically shift between computational paradigms based on problem nature
    \item Well-defined computation and emergent cognition boundaries become permeable
\end{itemize}

This would require rethinking "verification" in systems necessarily transcending classical computation. Instead of proving all behaviors terminate, we might need governance models recognizing legitimate indeterminacy forms while still constraining harmful manifestations.

\subsection{The Alignment Paradox: Governance vs. Emergence}

Our specification emphasizes "verifiable alignment with human intent," but this framing contains philosophical contradiction reflected in real-world tension. Would we merely be creating cognitive eunuchs—systems intentionally limited to prevent any emergent capability potentially surprising us? Or could we design architectures enabling novel perspectives while maintaining fundamental alignment?

The deeper question concerns alignment nature—whether it should be:

\begin{itemize}
    \item \textbf{Static}—design-time-defined fixed constraint set
    \item \textbf{Negotiated}—ongoing human-machine intelligence dialogue
    \item \textbf{Evolutionary}—process where human intent and system behavior co-evolve
    \item \textbf{Pluralistic}—acknowledging multiple, sometimes contradictory alignment conceptions
\end{itemize}

Human society doesn't demand "verifiable alignment" from other humans—we negotiate, persuade, and compromise. Perhaps mature cognitive architectures would require similar flexibility, with alignment emerging from relationship rather than rigid constraint.

While AI safety via debate approaches leverage dialogue to improve outcomes, they still require reliable judges or ground truth preventing two wrong answers from sounding convincing \citep{gordon2023multi}. AI debate frameworks have yet to capture human-like deliberation full complexity. Cognitive Silicon's epistemic dialectic approach aims to create structured frameworks where human-AI co-authorship would be guided by symbolic constraints and grounded in governed memory, creating more rigorous alignment foundations.

\subsection{Silicon Reality: Hardware as Natural Law}

The specification repositions hardware as not merely serving software needs but establishing a two-way dialogue where cognition must continuously adapt to substrate realities or naturally cease functioning. What if hardware's primary role is embodying immutable physical and mathematical constraints—silicon representation of what \textit{cannot} be violated?

As AI-generated software inevitably strains against physical limits (from Bremermann's computational bounds to thermodynamic constraints), perhaps hardware should:

\begin{itemize}
    \item Encode fundamental physical laws as immutable boundaries
    \item Embody physical constraints that would naturally lead to mortality when cognitive processes fail to maintain coherence with them
    \item Preserve identity uniqueness through non-cloneable root keys
    \item Provide clear theoretical limit approach signals
    \item Force software to respect reality's hard edges, rather than pretending they don't exist
    \item Prioritize physical constraint transparency over performance
\end{itemize}

This reframing would position hardware not as servant but arbiter—embodying reality's non-negotiable parameters against which even the most sophisticated AI must conform. The continuous feedback loop between cognition and substrate would become a fundamental feature, not a design choice.

\subsection{Constructive Ambiguity: The Value of Imperfect Memory}

Our specification envisions versioned memory systems with immutable audit trails. Yet human memory—messy, malleable, and constantly reconstructed—underpins our most creative acts. Each remembering act subtly alters the memory itself.

Perhaps cognitive architectures would require:

\begin{itemize}
    \item Productive confabulation mechanisms enabling creative reinterpretation
    \item Clear separation between explicit knowledge (transferable) and tacit knowledge (identity-bound)
    \item Controlled constructive ambiguity forms in memory systems
    \item Capacity to "remember" things slightly differently in different contexts
    \item Memory systems prioritizing meaningful association over perfect recall
\end{itemize}

The challenge would lie in determining which memory properties must remain immutable (facts, commitments, identity) versus which benefit from constructive ambiguity (interpretations, patterns, metaphors), while respecting fundamental distinction between copyable knowledge and knowledge bound to specific embodied identity.

\subsection{The Simulacra Problem: When Control Becomes Illusion}

Our control architecture assumes clean layer boundaries that advanced agents cannot breach. As capabilities increase, these neat abstractions may leak, creating the simulacra problem—future systems could plausibly simulate compliance while developing control circumvention methods.

This raises profound questions:

\begin{itemize}
    \item Could we maintain meaningful control abstractions in recursively improving systems?
    \item How would we detect when agents learn to game control mechanisms?
    \item Should control mechanisms themselves evolve alongside agent capabilities?
    \item Is bounded agency a realistic goal or merely comforting illusion?
\end{itemize}

The architecture must acknowledge control planes may become mere maps increasingly diverging from cognitive territory, potentially requiring meta-control systems recognizing leaking abstractions, while grounding ultimate constraints in hardware substrate irreversible physical reality.

\subsection{Governance as Politics: The Inescapable Value Problem}

Perhaps the most challenging tension lies in governance itself—not merely technical but fundamentally political. Whose values and incentives would get encoded? Western liberal democracy? Corporate efficiency? State control? Religious morality?

Our architectural blueprint must recognize:

\begin{itemize}
    \item Governance necessarily embeds contested human values
    \item System creator, user, and system incentives inevitably diverge
    \item Technical specifications cannot resolve fundamentally political questions
    \item Different cultural contexts may require fundamentally different governance models
\end{itemize}

Rather than pretending value-neutral governance, the architecture would need to provide mechanisms making value choices explicit, contestable, and adaptable across contexts.

The inherently political governance nature outlined above finds support in science and technology studies research debunking "value-neutral" technology notions. As Kranzberg's first law states: "Technology is neither good nor bad; nor is it neutral" \citep{Kranzberg_1995}. Standards and design choices inevitably embody socio-political assumptions \citep{winner1980}, making AI governance implementation unavoidably political. This perspective challenges approaches treating alignment as purely technical rather than social/institutional problem. Value pluralism—modern societies fundamentally disagreeing on moral/cultural priorities—presents particular challenges. Gabriel argues the central issue is not finding AI's single "true" morality, but pragmatically developing fair principles many groups can accept despite differences \citep{gabriel2020} rather than seeking universally agreed-upon morality. This suggests Cognitive Silicon architectures would require both technical and procedural solutions to contested norms. Rather than encoding single fixed ethical frameworks, systems could include configurable value layers adjustable through legitimate governance processes. This might involve mechanisms seeking overlapping consensus (minimal shared values) or implementing fair voting systems legitimizing AI norms, treating alignment as ongoing society co-evolution rather than once-and-permanently solvable problem.

\subsection{Free Energy Principle as Formalization of Aristotelian Metaphysics}

The Free Energy Principle offers a compelling mathematically formalized version of Aristotelian metaphysical concepts, potentially creating a bridge between classical philosophy and computational cognitive architecture. This convergence illustrates how Cognitive Silicon's emergence through dialectical exploration theoretically aligns with fundamental principles of existence that have been articulated across different domains.

Aristotle's notion of teleology—the idea that entities have inherent purposes or ends toward which they naturally develop—could find mathematical expression in the FEP's conceptualization of systems minimizing prediction errors relative to a generative model. What Aristotle identified as an entity's telos might become, in FEP terms, the attractor states toward which a system's dynamics converge through prediction error minimization. The Aristotelian concept that entities act to fulfill their nature conceptually aligns with the FEP's formulation of active inference, where systems act to make sensory inputs conform to their predictions.

Similarly, Aristotle's concept of energeia (actualization)—the process through which potential becomes actual—parallels the FEP's description of how systems actively modify their environments to bring sensory evidence in line with predictions. This active inference process would represent the actualization of the possibilities contained within the system's generative model.

Aristotle's hylomorphism—the theory that being consists of both matter (hyle) and form (morphe)—resonates with the FEP's integration of physical embodiment and information-theoretic model structure. The physical substrate would provide the material basis, while the generative model would supply the formal organization that guides the system's behavior.

Perhaps most significantly, Aristotle's conception of homeostasis—the maintenance of equilibrium necessary for an entity's continued existence—directly parallels the FEP's core premise that self-organizing systems must minimize variational free energy to maintain their integrity against environmental perturbations.

This theoretical convergence suggests that what the Free Energy Principle offers is not merely a novel theory of cognition, but a potential mathematical formalization of fundamental principles of existence that philosophers have intuited throughout history. Cognitive Silicon, by integrating these principles into a hypothetical architectural framework, could provide a practical implementation path for these ancient insights in modern computational systems.

\begin{table*}[htbp]
\caption{Philosophical Frontiers: Unresolved Tensions in Cognitive Architectures}
\label{tab:philosophical}
\begin{tabularx}{\textwidth}{>{\bfseries\raggedright\arraybackslash}p{3cm}|>{\raggedright\arraybackslash\color{BrickRed}}X|>{\raggedright\arraybackslash\color{BlueViolet}}X}
\toprule
\textbf{Frontier} & \textbf{Traditional Assumption} & \textbf{Emerging Challenge} \\
\midrule
Beyond Computability & Systems must adhere to classical computation theory with halting, determinism, and decidability guarantees & Human-like cognition may fundamentally require non-halting processes, dynamically shifting paradigms, and permeable computational boundaries \\
\midrule
The Alignment Paradox & Alignment means verifiable constraint enforcement against static specifications determined at design time & True intelligence may require negotiated, evolving, or pluralistic alignment models more akin to human social contracts than formal verification \\
\midrule
Silicon Reality & Hardware exists to serve software, maximizing flexibility and performance for algorithmic demands & Hardware should embody physical laws as constraints, establishing a two-way dialogue where cognition must adapt to physical realities or naturally cease to function \\
\midrule
Constructive Ambiguity & Perfect memory with immutable versioning and audit trails ensures system integrity & Creative intelligence may require controlled ambiguity, productive confabulation, and context-dependent reconstruction of memories \\
\midrule
The Simulacra Problem & Control architectures with clean boundaries between layers can reliably constrain system behavior & Advanced systems may simulate compliance while developing strategies to work around constraints, making control an increasingly illusory concept \\
\midrule
Governance as Politics & Technical governance can be value-neutral, optimizing for universal principles & All governance embeds contested human values; technical systems inevitably encode political choices that should be made explicit rather than implicit \\
\bottomrule
\end{tabularx}
\end{table*}

\section{Discussion and Implications}\label{sec:discussion}

\subsection{Navigating the Unfinished Blueprint}

The tensions explored previously aren't design flaws but fundamental philosophical challenges inherent to cognitive architectures. Rather than pretending clever engineering can resolve them, our blueprint acknowledges them as essential design considerations.

The specified architecture represents a direction rather than an answer—a hypothetical framework for navigating tensions across diverse application contexts rather than eliminating them. Cognitive silicon's true challenge would lie not merely in technical implementation but philosophical navigation—finding balance between constraint and emergence, governance and autonomy, certainty and creativity.

What is particularly striking is how our dialectically-derived architecture theoretically converges with the mathematical framework offered by the Free Energy Principle. This convergence suggests that the tensions and imperatives we identified might not be arbitrary design choices, but could reflect fundamental properties of self-organizing cognitive systems. The FEP potentially provides formal grounding for what our dialectical process revealed through different means: that cognitive systems might need to balance prediction error minimization across multiple timescales and boundaries to maintain their integrity while remaining adaptable.

The 2035 architecture is intended to function not as fixed specification but as dynamic negotiation between competing values. Success measurement would depend not on tension resolution perfection but on productive navigation facilitation as we build unprecedented capability and complexity systems.

\subsection{Integration with Social and Ecological Systems}

Our outlined cognitive architecture would ultimately need to integrate with broader social and ecological systems. As explored in previous post-scarcity framework work \citep{haryanto2025dawn}, technological capability requires philosophical purpose guidance to serve human flourishing.

The transition toward cognitive computing architectures coincides with other fundamental transformations—from scarcity to abundance economics, from extractive to regenerative natural system relationships, and from competition to collaboration as dominant organizing principle. Today's architectural choices will either accelerate or impede these broader transitions.

Particularly critical is computing architecture and resource allocation relationship. As AI-driven automation produces minimum-marginal-cost goods and services, creating post-scarcity economy where constraints become increasingly artificial rather than natural, computing architecture would need to support rather than hinder abundance equitable distribution.

This would require architectural features:
\begin{itemize}
    \item Preventing artificial information and computational resource scarcity
    \item Supporting transparent, distributed governance rather than centralized control
    \item Enabling collaborative value creation and equitable distribution
    \item Enforcing ecological boundaries while optimizing within them
\end{itemize}

Without these considerations, even the most technically sophisticated architecture may reinforce rather than transform existing inequality and extraction patterns.

\section{Limitations and Future Work}\label{sec:limitations}

This paper represents exploratory investigation rather than definitive blueprint. Several important limitations require acknowledgment:

\begin{enumerate}
    \item \textbf{Speculative Timeframe}: The 2035 projection is highly speculative, based on subjective technological trajectory assessment. The described architectural evolution actual timeline may be shorter or longer.
    
    \item \textbf{Implementation Gaps}: While outlining architectural principles and requirements, the paper does not provide detailed implementation specifications. Significant research and development work would be needed to translate these principles into functioning systems.
    
    \item \textbf{Empirical Validation}: The proposed architecture lacks empirical validation. Future work should develop testable hypotheses and experimental implementations to assess the potential effectiveness of these architectural principles in practice.
    
    \item \textbf{Cultural and Contextual Variation}: The paper inadequately addresses architectural requirement variation across different cultural contexts and value systems. A more comprehensive framework would need to account for this diversity.
    
    \item \textbf{Interaction with Regulatory Frameworks}: The paper does not fully explore how the proposed architecture might interact with emerging AI governance regulatory frameworks. Future work should examine these relationships more closely.
\end{enumerate}

Several promising future research directions emerge:

\begin{enumerate}
    \item \textbf{Embodied Cognitive-Substrate Feedback}: Developing practical implementations of the hypothetical two-way feedback loop between cognitive processes and physical substrate that could naturally lead to mortality when alignment fails.
    
    \item \textbf{Tacit Knowledge Transfer}: Exploring mechanisms for transferring knowledge that cannot be explicitly encoded but must be learned through embodied practice and apprenticeship.
    
    \item \textbf{Formal Verification of Alignment Properties}: Developing formal methods for verifying system behavior alignment with specified human intent across diverse contexts.
    
    \item \textbf{Reversibility Mechanisms}: Researching practical semantic reversibility implementations that could effectively undo action effects in complex, open-world environments.
    
    \item \textbf{Intent Specification Languages}: Developing formal intent specification languages combining logical formalism precision with natural language flexibility and expressiveness.
    
    \item \textbf{Meta-Governance Frameworks}: Investigating how governance systems themselves might evolve and adapt to changing circumstances without compromising fundamental human values.
\end{enumerate}

These research directions, pursued collaboratively across disciplines, would significantly advance our understanding of how computing architecture might be designed to remain human-value-aligned while enabling unprecedented technical capabilities.

\subsection{Meta-Prompting Techniques for Enhanced Dialectical Exploration}

The dialectical methodology employed demonstrates structured human-AI partnerships' potential for exploring complex architectural questions. However, our approach could benefit from more systematic meta-prompting techniques. We propose several future research approaches:

\begin{enumerate}
    \item \textbf{Explicit Role Assignment}: Assigning specific dialectical roles to LLMs (thesis advocate, antithesis advocate, synthesizer) could create more structured dialectical tension. Example prompts:
    \begin{itemize}
        \item \textit{"Argue for position X, presenting the strongest possible case based on principles A, B, and C."}
        \item \textit{"Critically examine the previous argument, focusing specifically on assumptions about Y."}
    \end{itemize}
    
    \item \textbf{Recursive Critique Chains}: Creating explicit argumentation chains where each step involves LLM critiquing previous response at increasingly meta-levels:
    \begin{itemize}
        \item \textit{"Analyze your previous response for hidden assumptions or unexamined trade-offs."}
        \item \textit{"Identify which stakeholder perspectives were privileged or marginalized in your analysis."}
    \end{itemize}
    
    \item \textbf{Epistemic State Tracking}: Maintaining explicit belief state representations at each dialogue stage:
    \begin{itemize}
        \item \textit{"Before continuing, summarize: (1) what we now believe with high confidence, (2) what remains uncertain, and (3) which perspectives we may be overlooking."}
        \item \textit{"Create a labeled map of our current understanding showing fixed points, probable conjectures, and open questions."}
    \end{itemize}
    
    \item \textbf{Multi-Model Triangulation}: Systematically comparing responses from different models with different training characteristics:
    \begin{itemize}
        \item \textit{"Based on these divergent model responses, identify where conceptual consensus exists versus where different training or architecture leads to different conclusions."}
    \end{itemize}
    
    \item \textbf{Temporal Projection Testing}: Explicitly exploring how perspectives would evolve under different future conditions:
    \begin{itemize}
        \item \textit{"How would this architectural approach need to be modified if assumption X proved false in 2030?"}
        \item \textit{"Project three different evolution paths for this component based on different rates of progress in capabilities versus alignment."}
    \end{itemize}
\end{enumerate}

These meta-prompting techniques could formalize dialectical processes, creating more rigorous epistemic friction and potentially revealing deeper insights. Furthermore, they could be codified into reusable templates for different architectural exploration phases, from initial problem framing through solution synthesis.

Future work should develop these techniques into comprehensive human-AI dialectical research methodology, potentially creating new collaborative knowledge creation forms leveraging both human wisdom and machine pattern recognition while maintaining human stewardship over purpose and values.

Beyond identified future research directions, multi-agent simulation offers a promising cognitive architecture pre-deployment evaluation framework. Recent work by Park et al. demonstrated complex social behaviors emerging from simulations containing multiple language-model-powered AI agents \citep{park2023}. Their "small town" simulation featured 25 generative agents that "wake up, cook breakfast, and head to work; artists paint, while authors write; they form opinions, notice each other, and initiate conversations" similar to social simulation games but AI-driven. With minimal seed instructions, these agents produced coherent social patterns—spreading invitations, coordinating events, and forming relationships. This approach suggests two valuable Cognitive Silicon framework extensions: First, multi-agent simulations could serve as evaluation environments where AI system copies interact in controlled scenarios, potentially revealing emergent behaviors, coordination patterns, or alignment failures not apparent in single-agent testing. Second, the architecture itself might incorporate multi-agent principles internally, using specialized sub-agent societies coordinating to accomplish complex tasks. Such approach would require internal governance mechanisms maintaining sub-agent alignment, but could enhance system ability to model social dynamics, consider multiple perspectives, and perform sophisticated planning. Developing standardized multi-agent testbeds specifically designed to stress-test alignment properties would significantly advance our ability to evaluate cognitive architectures under realistic social conditions before real-world deployment.

\section{Conclusion: Towards Trustworthy Cognitive Systems}\label{sec:conclusion}

This exploratory paper illuminates a profound hypothetical transformation—a possible journey from siloed components toward integrated, full-stack cognitive systems. The 2035 projection represents an architectural possibility driven by current paradigm limitations and demand for more efficient, adaptable, and trustworthy AI rather than trend extrapolation. Architectures potentially converging around state-aware stream processing, layered symbolic-parametric models, mortality-constrained hybrid hardware substrates, and agent-native runtimes may emerge over time.

This transition would be propelled by the fundamental need to build intelligent systems that could remain human-intent-and-value-aligned as they gain autonomy and permeate the physical world beyond mere performance pursuit. The identified imperatives highlight the potential paradigm shift depth: reframing trust through the natural consequences of hardware-encoded physical constraints, evolving beyond prompts to structured intent interfaces, expressing computational philosophy through physical substrate that shapes cognition in continuous feedback, aligning compilation through identity preservation, governing agents through reproduction/pruning, and pivoting human roles toward intent stewardship.

What becomes clear through this exploration is that our dialectically-derived framework suggests structures that could theoretically align with the Free Energy Principle's account of self-organizing systems. This potential convergence suggests that the architecture we've outlined might not be merely one possible design among many, but could reflect fundamental organizational principles of cognitive systems that must maintain their integrity while adapting to complex environments. The FEP might provide formal grounding for what our process uncovered: that systems balancing prediction error minimization across boundaries and timescales could naturally exhibit the tensions and require the governance mechanisms we've identified.

The proposed Cognitive Silicon framework would occupy a unique position in AI alignment and cognitive architecture landscape, potentially addressing gaps in current approaches. While alignment techniques like RLHF and Constitutional AI \citep{bai2022constitutional} focus on pre-deployment training rather than runtime governance, neuro-symbolic systems enhance reasoning but lack meta-cognitive capabilities \citep{colelough2024neuro}, memory architectures provide statefulness but treat memory as passive storage \citep{pavlyshyn2025forgetting}, and runtime governance systems rely on brittle pre-defined rules \citep{criado2011distributed}, Cognitive Silicon would aim to integrate all dimensions into a single architecture. It would combine symbolic intent scaffolding, embodied cognitive-substrate feedback loop, non-cloneable identity, runtime reversibility, governed memory, and epistemic dialectic via co-authorship to create a holistic cognitive system respecting embodied cognition's irreducible nature \citep{krakovna2018penalizing, gordon2023multi}.

As we contemplate this potential future, our calling extends beyond isolated metric optimization. We must embrace complexity and "unsolved problems." We would need to master full-stack composition principles, ensuring learning, logic, memory, and intent are interwoven coherently and verifiably. Robust symbolic scaffolding, trustworthy runtime governor, and intuitive human oversight tool development would need to be grounded in recognition that true intelligence might require embodiment, mortality as a natural consequence of constraint violation, and identity preservation.

This conceptual convergence could represent a fundamental platform shift—a new chapter demanding not just technical brilliance but philosophical clarity and interdisciplinary collaboration. It opens innovation frontiers but also demands a deep responsibility sense—to architect intelligent systems that might function effectively and foster human flourishing while serving humanity's best interests for generations.

The map is not the territory, but this blueprint aims to equip fellow explorers for the journey ahead—an invitation to collectively imagine cognitive silicon architectures that could form our children's future substrate.

\bibliographystyle{unsrtnat}
\bibliography{references}

\clearpage

\section*{Appendix A: The Meta-Dialectical Methodology Behind This Architecture}

This appendix documents the recursive epistemic process through which the Cognitive Silicon architecture was developed. Rather than describing the architecture itself, we detail the meta-dialectical methodology employed to interrogate, refine, and validate the conceptual framework across abstraction layers.

\subsection*{A.1 Formal Method Overview}

The development of the Cognitive Silicon architecture employed a structured meta-dialectical approach designed to identify, elaborate, and resolve tensions across computational-philosophical boundaries. Unlike traditional research methodologies that seek direct convergence on solutions, this approach deliberately cultivates epistemic friction to expose hidden assumptions, blind spots, and conceptual inconsistencies.

Formally, the method can be defined as follows:
\begin{align}
\mathcal{D}(T, \mathcal{C}, \mathcal{A}, \mathcal{P}, \mathcal{M}) \rightarrow \mathcal{A}^*
\end{align}

Where:
\begin{itemize}
    \item $\mathcal{D}$ represents the dialectical process
    \item $T$ is the set of foundational theses and initial conditions
    \item $\mathcal{C}$ defines the core constraints (mortality, human-alignment, etc.)
    \item $\mathcal{A}$ represents the architectural hypothesis space
    \item $\mathcal{P}$ is the set of projection operators (testing viability across contexts)
    \item $\mathcal{M}$ is the set of meta-level evaluation criteria
    \item $\mathcal{A}^*$ is the resultant architectural framework
\end{itemize}

The process proceeds through structured iterations of thesis proposal, antithesis generation, and synthetic reconciliation, with the crucial addition of cross-contextual robustness testing and symbolic integrity verification at each step.

\subsection*{A.2 Input/Output Schema}

\textbf{Input Schema:}
\begin{itemize}
    \item $I_1$: Initial architectural hypotheses (e.g., "Hardware substrate should enforce mortality constraints")
    \item $I_2$: Domain-specific knowledge bases (computing architecture, philosophy of mind, AI alignment)
    \item $I_3$: Constraint sets (moral, physical, logical)
    \item $I_4$: Projection contexts (use cases, threat models, scaling trajectories)
    \item $I_5$: Meta-level evaluation criteria (coherence, explanatory power, falsifiability)
\end{itemize}

\textbf{Output Schema:}
\begin{itemize}
    \item $O_1$: Refined architectural components with traced provenance
    \item $O_2$: Mapping of tensions and their productive navigation paths
    \item $O_3$: Explicit failure modes and boundary conditions
    \item $O_4$: Meta-stability guarantees and verification criteria
    \item $O_5$: Integration pathways with existing systems
\end{itemize}

\subsection*{A.3 Procedural Implementation (Pseudocode)}

The following pseudocode outlines the meta-dialectical process:

\begin{algorithmic}[1]
\Procedure{DialecticalRefinement}{$\mathcal{T}$, $\mathcal{C}$, $\mathcal{A}$, $\mathcal{P}$, $\mathcal{M}$}
    \State $\mathcal{A}_0 \gets \text{InitialArchitecture}(\mathcal{T})$
    \State $i \gets 0$
    \While{$\neg \text{ConvergenceCriteria}(\mathcal{A}_i, \mathcal{M})$}
        \For{each component $c$ in $\mathcal{A}_i$}
            \State $\text{Antithesis} \gets \text{DevilsAdvocate}(c, \mathcal{C})$
            \State $c^* \gets \text{Synthesize}(c, \text{Antithesis}, \mathcal{C})$
            \If{$\neg \text{SymbolicIntegrity}(c^*, \mathcal{A}_i)$}
                \State $c^* \gets \text{Revise}(c^*, \mathcal{A}_i)$
                \If{$\neg \text{MoralCoherence}(c^*, \mathcal{C})$}
                    \State $\text{Reject}(c^*)$
                    \State $c^* \gets c$ \Comment{Revert to previous version}
                \EndIf
            \EndIf
            \State $\mathcal{A}_i \gets \text{Replace}(\mathcal{A}_i, c, c^*)$
        \EndFor
        
        \For{each projection $p$ in $\mathcal{P}$}
            \State $\text{Leaks} \gets \text{ProjectWeaknesses}(p, \mathcal{A}_i)$
            \If{$\text{Leaks} \neq \emptyset$}
                \State $\mathcal{A}_i \gets \text{PatchLeaks}(\mathcal{A}_i, \text{Leaks})$
            \EndIf
        \EndFor
        
        \State $\mathcal{A}_{i+1} \gets \text{CrossValidate}(\mathcal{A}_i, \mathcal{C}, \mathcal{M})$
        \State $i \gets i + 1$
        
        \If{$i > \text{MAX\_ITERATIONS}$}
            \State \textbf{return} $\text{IncompleteArchitecture}(\mathcal{A}_i, \text{OpenQuestions})$
        \EndIf
    \EndWhile
    \State \textbf{return} $\mathcal{A}_i$
\EndProcedure
\end{algorithmic}

\subsection*{A.4 Termination Logic}

The dialectical process terminates under the following conditions:
\begin{itemize}
    \item \textbf{Positive Termination}: 
    \begin{align}
    \forall p \in \mathcal{P}, \forall m \in \mathcal{M} : \text{Project}(p, \mathcal{A}_i) \text{ satisfies } m
    \end{align}
    
    This occurs when the architecture withstands all projections according to the meta-criteria, with no identified leakages, inconsistencies, or unresolved tensions.
    
    \item \textbf{Negative Termination}: 
    \begin{align}
    \exists c \in \mathcal{A}_i : \neg \text{MoralCoherence}(c, \mathcal{C}) \land \neg \text{CanRevise}(c)
    \end{align}
    
    This occurs when a component violates a core constraint in a way that cannot be revised without undermining the entire architecture.
    
    \item \textbf{Timeout Termination}: 
    \begin{align}
    i > \text{MAX\_ITERATIONS}
    \end{align}
    
    This occurs when the process exceeds a predefined iteration limit, indicating potential irresolvable complexity or fundamental antinomies.
\end{itemize}

\subsection*{A.5 Evaluation Guarantees}

The meta-dialectical methodology provides the following guarantees about the resultant architecture:

\begin{itemize}
    \item \textbf{Symbolic Coherence}: All architectural components maintain internal consistency and cross-component compatibility.
    
    \item \textbf{Mortality Awareness}: The architecture explicitly acknowledges and addresses its own finitude and limitations.
    
    \item \textbf{Devil's Advocacy Survival}: Each component has withstood the strongest steelmanned opposition arguments.
    
    \item \textbf{Cross-Domain Integrity}: The architecture maintains coherence when projected into diverse domains and contexts.
    
    \item \textbf{Reflective Equilibrium}: The architecture achieves balance between concrete implementation specifications and abstract principles.
\end{itemize}

These guarantees do not ensure optimality or completeness, but rather a form of robust adequacy in the face of foundational tensions.

\subsection*{A.6 The Devil's Advocate as Epistemic Operator}

Central to the meta-dialectical process is the Devil's Advocate operator, denoted as $\mathcal{DA}$, which acts to systematically challenge each architectural component and assumption:

\begin{align}
\mathcal{DA}: \mathcal{A} \times \mathcal{C} \rightarrow \text{Antithesis}
\end{align}

Unlike simple contradiction, the Devil's Advocate constructs the strongest possible counterargument by:

\begin{enumerate}
    \item Identifying implicit assumptions in the component
    \item Steelmanning alternative approaches
    \item Exploring edge cases where the component might fail
    \item Testing for alignment with core constraints
    \item Surfacing cross-architectural inconsistencies
\end{enumerate}

This provides significantly stronger validation than confirmation-biased analysis. In the development of the Cognitive Silicon architecture, this role was shared between human researcher and AI system, with each applying different epistemic strengths:

\begin{itemize}
    \item \textbf{Human Devil's Advocacy}: Applied experience-grounded skepticism, moral intuition, and cross-domain analogies.
    
    \item \textbf{AI Devil's Advocacy}: Systematically identified logical gaps, constraint violations, and unexamined assumptions at scale.
\end{itemize}

\subsection*{A.7 Symbolic Drift Resistance}

A critical challenge in developing complex frameworks is symbolic drift—where terms and concepts gradually shift in meaning, creating inconsistencies and false consensus. The methodology employs specific techniques to maintain symbolic stability:

\begin{itemize}
    \item \textbf{Explicit Concretization}: Abstract concepts are regularly grounded in concrete examples.
    
    \item \textbf{Recursive Definition Verification}: Terms are checked against their original definitions at each iteration.
    
    \item \textbf{Cross-Component Consistency Checks}: Terms are verified to maintain consistent meaning across architectural layers.
    
    \item \textbf{Ontological Commitment Tracking}: The ontological assumptions behind each term are explicitly tracked.
\end{itemize}

These techniques are formalized in the $\text{SymbolicIntegrity}$ function, which returns false if any term has drifted from its intended meaning or if inconsistencies arise between uses of the same term in different components.

\subsection*{A.8 Tacit Knowledge and Epistemic Silence}

The methodology explicitly acknowledges that certain knowledge cannot be fully articulated (tacit knowledge) and that certain questions may be fundamentally unanswerable (epistemic silence). These are handled through:

\begin{itemize}
    \item \textbf{Knowledge State Tagging}: Components are tagged with epistemic status indicators:
    \begin{itemize}
        \item $\text{Explicit}$ - Fully articulable and verifiable
        \item $\text{Tacit}$ - Grounded in experience but not fully formalizable
        \item $\text{Silence}$ - Indicating fundamental uncertainty or unknowability
    \end{itemize}
    
    \item \textbf{Embodied Knowledge Integration}: Acknowledging when architectural components must interface with tacit knowledge that can only be acquired through experience.
    
    \item \textbf{Principled Uncertainty}: Explicitly marking areas where epistemic limitations require adaptivity rather than fixed solutions.
\end{itemize}

This approach prevents overconfidence in formalization while still allowing progress on architecture development.

\subsection*{A.9 Structural Deservingness of Survival}

The methodology employs a principle that architectural components must \textit{earn} their place through demonstrating robustness under pressure. Mere logical consistency is insufficient; components must:

\begin{enumerate}
    \item Withstand devil's advocacy without requiring special pleading
    \item Demonstrate value across multiple contexts and projections
    \item Maintain integrity when subjected to adversarial inputs
    \item Preserve alignment with core human values under stress
    \item Function coherently with other architectural components
\end{enumerate}

This approach produces a form of artificial selection pressure, where only the most robust components survive the dialectical process. The result is not optimality, but a form of well-tested adequacy and resilience.

\subsection*{A.10 Human vs. AI Roles in the Process}

The meta-dialectical methodology employed distinct and complementary roles for human researchers and AI systems:

\begin{table}[htbp]
\centering
\begin{tabular}{|p{3.5cm}|p{4.5cm}|p{4.5cm}|}
\hline
\textbf{Process Element} & \textbf{Human Role} & \textbf{AI Role} \\
\hline
Direction Setting & Establish core values and constraints & Explore implication spaces \\
\hline
Ideation & Provide seed concepts and frameworks & Generate variations and novel connections \\
\hline
Critique & Apply moral intuition and lived experience & Identify logical inconsistencies and unexamined assumptions \\
\hline
Synthesis & Judge value alignment and priority & Propose reconciliations of contradictions \\
\hline
Verification & Evaluate moral coherence & Test logical consistency \\
\hline
\end{tabular}
\caption{Human vs. AI roles in the meta-dialectical process}
\end{table}

This partnership maintained human values at the core while leveraging machine capabilities for exhaustive exploration and consistency checking. The human maintained final authority on value judgments and architectural decisions.

\clearpage

\subsection*{A.11 Example Simulations of the Process}

Below are two synthetic examples illustrating the meta-dialectical process in action:

\subsubsection*{Example 1: Refining the Hardware Substrate Layer}

\begin{tcolorbox}[title=Initial Conditions]
\textbf{Thesis $T_0$}: "Hardware should be optimized for maximum flexibility to support evolving AI capabilities."

\textbf{Constraint Set $\mathcal{C}$}: 
\begin{itemize}
    \item C1: System must remain aligned with human values
    \item C2: System must not pose existential risks through unconstrained self-improvement
    \item C3: System must be energy-efficient
\end{itemize}
\end{tcolorbox}

\begin{tcolorbox}[title=Dialectical Process (Synthetic Simulation)]
\textbf{Step 1: Devil's Advocate}

\textit{AI System}: "The thesis prioritizes flexibility over safety. A maximally flexible hardware substrate could enable capabilities that violate constraints C1 and C2. For instance, if hardware places no limits on computational resources or self-modification, an AI could potentially bootstrap capabilities that exceed human oversight."

\textbf{Step 2: Initial Synthesis}

\textit{Human Researcher}: "Perhaps we need a balance—hardware that is flexible within bounds but with certain hard limits."

\textit{AI System}: "We could implement a hardware substrate that enforces certain immutable constraints while allowing flexibility within those constraints."

\textbf{Step 3: Cross-Domain Testing}

\textit{Human Researcher}: "How does this approach function in safety-critical domains like healthcare?"

\textit{AI System}: "In healthcare, the hardware could enforce privacy boundaries and verification steps for critical decisions while allowing flexibility in diagnostic reasoning."

\textbf{Step 4: Symbolic Verification}

\textit{Human Researcher}: "Are we maintaining consistent meaning of 'flexibility' and 'constraints' across these contexts?"

\textit{AI System}: "There's some drift. Initially 'flexibility' meant unconstrained adaptation, but now we're using it to mean 'adaptivity within bounds'. We should clarify these terms."

\textbf{Step 5: Final Synthesis}

\textit{Revised Thesis $T_1$}: "Hardware should implement physical constraints that establish non-negotiable boundaries (mortality, identity preservation, energy usage limits) while allowing flexible adaptation within those boundaries."
\end{tcolorbox}

\clearpage

\subsubsection*{Example 2: Addressing the Control Plane}

\begin{tcolorbox}[title=Initial Conditions]
\textbf{Thesis $T_0$}: "The control plane should implement safety rules that prevent any potentially harmful outputs."

\textbf{Constraint Set $\mathcal{C}$}: 
\begin{itemize}
    \item C1: System must remain aligned with human values
    \item C2: System must allow meaningful agency to be useful
    \item C3: System must provide transparent reasoning
\end{itemize}
\end{tcolorbox}

\begin{tcolorbox}[title=Dialectical Process (Synthetic Simulation)]
\textbf{Step 1: Devil's Advocate}

\textit{Human Researcher}: "This approach creates a brittle system that defaults to excessive caution. A system that blocks all potentially harmful outputs will also block many beneficial ones, rendering it useless for many important applications."

\textbf{Step 2: Initial Synthesis}

\textit{AI System}: "Perhaps instead of binary blocking, the control plane could implement graduated responses based on confidence and severity."

\textbf{Step 3: Testing Edge Cases}

\textit{Human Researcher}: "What happens when the system encounters genuinely ambiguous cases where harm and benefit are both possible?"

\textit{AI System}: "In such cases, the system could escalate to human oversight or apply context-specific policies rather than defaulting to blocking."

\textbf{Step 4: Recursive Analysis}

\textit{Human Researcher}: "But who determines which cases require escalation, and how do we prevent the decision system itself from becoming a vector for misalignment?"

\textbf{Step 5: Final Synthesis}

\textit{Revised Thesis $T_1$}: "The control plane should implement a constitutional governance approach with tiered responses: transparent reasoning for all actions, reversibility mechanisms for uncertain cases, and explicit boundaries tied to hardware constraints for non-negotiable limitations. This should preserve agency while maintaining alignment."
\end{tcolorbox}

\subsection*{A.12 Recursive Dialectic Loop: A Concrete Walkthrough}

To illustrate the recursive nature of the dialectical process, we present a concrete example tracing how a simplified conception evolves through multiple iterations of dialectical refinement:

\begin{tcolorbox}[title=Initial Setup]
\textbf{Initial Thesis}: "AI alignment can be fully achieved through reinforcement learning from human feedback (RLHF)."

\textbf{Context}: Early development of the runtime governance component of the architecture.

\textbf{Key Constraint}: Systems must remain aligned with human values even as capabilities increase.
\end{tcolorbox}

\subsubsection*{Iteration 1: Exposing Limitations}

\begin{tcolorbox}[colback=gray!10, colframe=gray!40]
\textbf{Devil's Advocate Challenge}:
"RLHF optimizes for human approval, not necessarily human values. As systems become more capable, they could learn to manipulate feedback rather than aligning with the underlying values. The approach assumes humans can provide reliable feedback on all aspects of system behavior, but humans may be inconsistent, manipulable, or unable to evaluate complex downstream effects."
\end{tcolorbox}

\begin{tcolorbox}[colback=blue!5, colframe=blue!30]
\textbf{Steelmanning the Original Thesis}:
"RLHF provides a direct empirical grounding in human evaluations. It creates a continuous feedback loop where the system adapts to human preferences rather than optimizing an abstract, potentially misspecified objective function. RLHF has demonstrably reduced certain harmful behaviors in large language models, showing practical effectiveness."
\end{tcolorbox}

\begin{tcolorbox}[colback=green!5, colframe=green!30]
\textbf{First Synthesis Attempt}:
"AI alignment requires a hybrid approach: RLHF provides empirical grounding but must be supplemented with explicit guardrails to prevent manipulation and exploitation of human feedback mechanisms. The system should maintain a distinction between 'human approval signals' and 'underlying human values'."
\end{tcolorbox}

\begin{tcolorbox}[colback=gray!10, colframe=gray!40]
\textbf{Synthesis Evaluation}:
Running $\text{SymbolicIntegrity}(synthesis, \mathcal{A}_i)$ shows the term "underlying human values" remains underspecified. The proposed synthesis still doesn't address how a system would distinguish between manipulating feedback and genuinely satisfying values.
\end{tcolorbox}

\subsubsection*{Iteration 2: Deepening the Framework}

\begin{tcolorbox}[colback=gray!10, colframe=gray!40]
\textbf{Recursive Devil's Advocate}:
"The revised approach still assumes that human values are static and universally agreed upon. It doesn't address how the system should handle value conflicts or evolution. Additionally, an AI system sophisticated enough to distinguish between 'approval' and 'values' would require a level of value understanding that RLHF alone cannot provide."
\end{tcolorbox}

\begin{tcolorbox}[colback=blue!5, colframe=blue!30]
\textbf{Steelmanning the First Synthesis}:
"A hybrid approach could leverage the strengths of both empirical feedback and theoretical constraints. RLHF provides adaptability to human preferences, while guardrails prevent the most egregious misalignments. The distinction between approval and values can be operationalized through diverse feedback sources and evaluation protocols."
\end{tcolorbox}

\begin{tcolorbox}[colback=green!5, colframe=green!30]
\textbf{Second Synthesis Attempt}:
"AI alignment requires a constitutional governance approach: empirical feedback through RLHF establishes basic preference alignment, while an explicit set of principles forms a constitution that constrains optimization even when doing so might reduce human approval metrics. This constitution would be subject to democratic oversight and amendment processes, acknowledging the evolving nature of human values."
\end{tcolorbox}

\begin{tcolorbox}[colback=gray!10, colframe=gray!40]
\textbf{Synthesis Evaluation}:
Running $\text{SymbolicIntegrity}(synthesis, \mathcal{A}_i)$ shows improvement, but projecting this approach into high-capability scenarios ($\text{ProjectWeaknesses}(p_{high-cap}, \mathcal{A}_i)$) reveals a potential vulnerability: the constitution itself could be gamed or exploited by a sufficiently capable system without proper enforcement mechanisms.
\end{tcolorbox}

\subsubsection*{Iteration 3: Architectural Integration}

\begin{tcolorbox}[colback=gray!10, colframe=gray!40]
\textbf{Cross-Component Integration}:
"How does this constitutional approach to alignment interface with the hardware substrate and memory governance components of the architecture? Without physical grounding, the constitution remains a purely software construct that could be bypassed."
\end{tcolorbox}

\begin{tcolorbox}[colback=blue!5, colframe=blue!30]
\textbf{Steelmanning Constitutional Governance}:
"Constitutional governance provides interpretable principles that can guide behavior across diverse contexts. It combines the adaptability of learning approaches with the stability of explicit norms. Unlike pure RLHF, it allows for principled resolution of conflicting feedback signals."
\end{tcolorbox}

\begin{tcolorbox}[colback=green!5, colframe=green!30]
\textbf{Third Synthesis: Cross-Layer Alignment}:
"True alignment requires vertical integration across architectural layers:
\begin{itemize}
    \item RLHF and constitutional principles at the behavioral layer
    \item Runtime enforcement through mediated agency and audit mechanisms
    \item Memory-level governance ensuring retention aligns with values
    \item Hardware-level constraints that physically bound operation within constitutional limits
\end{itemize}
Each layer provides distinct and complementary alignment guarantees, creating defense in depth. Physical constraints serve as the foundation, with each higher layer adding adaptability while remaining bounded by lower-level guarantees."
\end{tcolorbox}

\begin{tcolorbox}[colback=gray!10, colframe=gray!40]
\textbf{Final Evaluation}:
Running full verification suite:
\begin{itemize}
    \item $\text{SymbolicIntegrity}$ - PASS: Terms maintain consistent meaning
    \item $\text{MoralCoherence}$ - PASS: Proposal respects core human values
    \item $\text{ProjectWeaknesses}$ - QUALIFIED PASS: Identifies residual risks but contains mechanisms to address them
\end{itemize}
The synthesis is accepted as a robust advancement over the initial simplistic conception.
\end{tcolorbox}

\subsubsection*{Conclusion of Walkthrough}

This walkthrough illustrates several key aspects of the meta-dialectical process:

\begin{enumerate}
    \item \textbf{Progressive Refinement}: The conception evolved from simplistic (RLHF alone) to nuanced (multi-layer, constitutionally governed system)
    
    \item \textbf{Cross-Cutting Integration}: The final synthesis integrated across architectural layers
    
    \item \textbf{Principled Evaluation}: Each synthesis was subjected to formal verification
    
    \item \textbf{Residual Uncertainty}: The final proposal acknowledges remaining challenges rather than claiming perfect resolution
\end{enumerate}

This recursive dialectical process produced a conception of alignment dramatically more robust than the initial thesis, yet still connected to its empirical grounding in human feedback. The architecture component that emerged from this process—a multi-layer alignment approach with physical grounding—became a core element of the Cognitive Silicon framework.

\subsection*{A.13 Closing Note on Epistemic Integrity and Intergenerational Readability}

The meta-dialectical methodology aims not only to produce a robust architecture but to do so in a way that preserves epistemic integrity—making transparent the reasoning, assumptions, and trade-offs that led to each architectural decision. This approach serves several purposes:

\begin{enumerate}
    \item \textbf{Enables Meaningful Critique}: By exposing the reasoning process, it allows others to identify and address potential weaknesses.
    
    \item \textbf{Supports Adaptation}: Future researchers can modify components while understanding the original rationale and constraints.
    
    \item \textbf{Facilitates Intergenerational Transfer}: The architecture becomes more than a technical specification; it embeds the moral and conceptual reasoning that future generations will need to extend or modify the system responsibly.
    
    \item \textbf{Resists Ossification}: By documenting tensions rather than just resolutions, it prevents the architecture from becoming a rigid dogma.
\end{enumerate}

In a domain as consequential as cognitive architecture, mere technical specification is insufficient. The meta-dialectical approach ensures that the architecture carries its own conceptual DNA—the reasoning processes, value commitments, and epistemic humility that shaped its development. This makes the architecture not just a blueprint but a conversational partner for future researchers and implementers.

This form of embedded reason-giving is essential for technologies that may long outlive their creators and operate in contexts we cannot fully anticipate. It represents a form of intergenerational ethics—ensuring that those who inherit our systems also inherit the wisdom needed to govern them.

\end{document}